\documentclass[10pt,twocolumn]{article}

\usepackage[T1]{fontenc}
\usepackage{lmodern}
\usepackage[margin=0.72in]{geometry}
\usepackage{booktabs}
\usepackage{graphicx}
\usepackage{microtype}
\usepackage{tabularx}
\usepackage{tikz}
\usepackage{xcolor}
\usepackage{hyperref}
\usepackage{url}
\usetikzlibrary{arrows.meta,positioning}

\hypersetup{
  colorlinks=true,
  linkcolor=blue!55!black,
  citecolor=blue!55!black,
  urlcolor=blue!55!black
}

\newcommand{\system}{Meganeura}
\newcommand{\inferena}{Inferena}
\newcommand{\code}[1]{\texttt{#1}}

\title{\system{}: Portable GPU Training and Inference\\
through Vulkan and Metal}
\author{Dzmitry Malyshau\\[-0.1em]
{\small Independent Researcher}\\[-0.1em]
{\small \href{mailto:kvark@fastmail.com}{kvark@fastmail.com}
\quad
\href{https://orcid.org/0009-0005-6410-4276}{ORCID: 0009-0005-6410-4276}}\\[-0.1em]
{\small Project and artifact: \url{https://github.com/kvark/meganeura}}}
\date{}

\begin{document}
\maketitle

\begin{abstract}
Training and deployed inference often cross export, conversion, and
platform-specific runtime boundaries. \system{} asks whether one compact
native compiler can span both phases on consumer GPUs. Its typed static graph,
automatic differentiation, optimizer, checkpoint, memory planner, and runtime
lower specialized programs through Vulkan and Metal.

We compare five matched workloads with PyTorch on NVIDIA and AMD discrete
GPUs, an AMD APU, Apple silicon, and an Intel iGPU. The protocol separates
strict f32 from validated fast paths and gates forward and backward
independently. Forty-eight of 50 device--workload--mode cells pass both gates;
the other two share one unresolved backward-reference disagreement on a newly
supported APU. In strict f32, \system{} wins 12 of 20 GPU-referenced
minimal-latency cells and has a median valid training gap of $1.8\times$. On
the discrete AMD GPU, four of five inference workloads are within
$1.10\times$ of compiled ROCm PyTorch and three training workloads are faster.
Under accelerated contracts, the worst training gap is $4.6\times$.

Compilation takes 0.1--2.4\,s versus 6--96\,s for \code{torch.compile} on
supported GPU paths; the stripped binary is 13\,MiB. Dispatch profiles
localize the largest gaps to convolution derivatives and attention backward.
A physical Android XR case study transfers a \system{}-trained decoder into
an Adreno/OpenXR application sharing the graphics queue. The results show
that general consumer graphics APIs can support a compact shared
train-to-deploy stack at useful, sometimes vendor-competitive performance.
The measured gaps point to kernel coverage, scheduling, and arithmetic policy
rather than an identified API limitation.
\end{abstract}

\section{Introduction}

The software path used to train a model is often not the path used to deploy
it. Training commonly relies on PyTorch~\cite{paszke2019pytorch} and vendor
libraries; deployment may use TensorRT, Core ML, a mobile runtime, a browser
engine, or a bespoke application stack. Each boundary duplicates model
conversion, operator coverage, numerical policy, profiling, and correctness
work. It also makes adaptation on the deployed device unusually difficult.

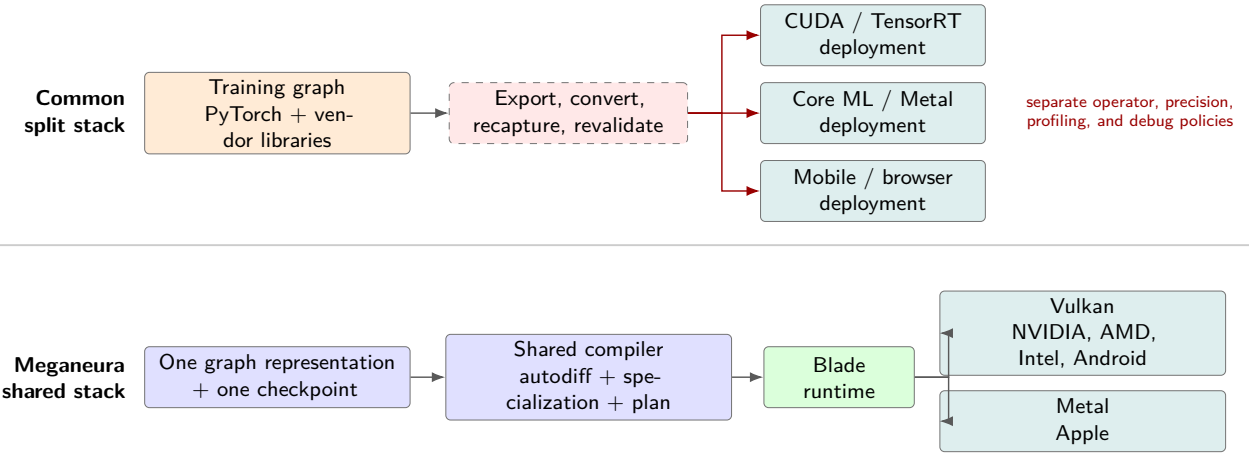
\begin{figure*}[t]
\centering
\resizebox{0.97\textwidth}{!}{%
\begin{tikzpicture}[
  font=\sffamily\small,
  box/.style={
    draw=black!55,
    rounded corners=2pt,
    minimum height=9mm,
    align=center,
    inner xsep=6pt,
    inner ysep=3pt
  },
  lane/.style={font=\sffamily\bfseries\small, align=right},
  train/.style={box, fill=orange!16},
  boundary/.style={box, fill=red!9, dashed},
  shared/.style={box, fill=blue!12},
  runtime/.style={box, fill=green!14},
  target/.style={box, fill=teal!13},
  arrow/.style={-{Latex[length=2mm]}, semithick, draw=black!65},
  split/.style={-{Latex[length=2mm]}, semithick, draw=red!60!black}
]
  \node[lane, text width=23mm] (split-label) at (0,1.2)
    {Common\\split stack};
  \node[train, text width=35mm] (pytorch) at (3.4,1.2)
    {Training graph\\PyTorch + vendor libraries};
  \node[boundary, text width=31mm] (convert) at (7.7,1.2)
    {Export, convert,\\recapture, revalidate};
  \node[target, text width=29mm] (tensorrt) at (12.2,2.35)
    {CUDA / TensorRT\\deployment};
  \node[target, text width=29mm] (coreml) at (12.2,1.2)
    {Core ML / Metal\\deployment};
  \node[target, text width=29mm] (mobile) at (12.2,0.05)
    {Mobile / browser\\deployment};
  \node[align=center, text=red!60!black, font=\sffamily\scriptsize]
    at (16.0,1.2)
    {separate operator, precision,\\profiling, and debug policies};

  \draw[arrow] (pytorch) -- (convert);
  \draw[split] (convert.east) -- ++(5mm,0) |- (tensorrt.west);
  \draw[split] (convert.east) -- (coreml.west);
  \draw[split] (convert.east) -- ++(5mm,0) |- (mobile.west);

  \draw[black!20, thick] (-1.4,-0.75) -- (18.0,-0.75);

  \node[lane, text width=23mm] (one-label) at (0,-2.7)
    {\system{}\\shared stack};
  \node[shared, text width=35mm] (graph) at (3.4,-2.7)
    {One graph representation\\+ one checkpoint};
  \node[shared, text width=38mm] (compiler) at (8.0,-2.7)
    {Shared compiler\\autodiff + specialization + plan};
  \node[runtime, text width=18mm] (blade) at (11.7,-2.7)
    {Blade\\runtime};
  \node[target, text width=38mm] (vulkan) at (15.3,-2.05)
    {Vulkan\\NVIDIA, AMD, Intel, Android};
  \node[target, text width=38mm] (metal) at (15.3,-3.35)
    {Metal\\Apple};

  \draw[arrow] (graph) -- (compiler);
  \draw[arrow] (compiler) -- (blade);
  \draw[arrow] (blade.east) -- ++(5mm,0) |- (vulkan.west);
  \draw[arrow] (blade.east) -- ++(5mm,0) |- (metal.west);
\end{tikzpicture}%
}
\caption{The systems question. A conventional train/deploy boundary can split
one model into several runtime-specific paths. \system{} studies the
alternative: keep graph semantics, automatic differentiation, specialization,
checkpointing, and execution in one native stack while changing only the
graphics backend. The top lane is illustrative, not a feature-equivalence
claim about the named products.}
\label{fig:fragmentation}
\end{figure*}

Consumer graphics APIs offer a tempting common substrate for the alternative
in Figure~\ref{fig:fragmentation}. Vulkan covers desktop, mobile, and embedded
GPUs across vendors, while Metal exposes Apple GPUs. Both offer general
compute and increasingly expose matrix hardware. They do not, however,
provide the complete compiler, kernel library, graph-capture, and debugging
ecosystem available to CUDA applications. A portable system must decide which
layers it can replace, which hardware features it can use safely, and how much
performance it loses.

This paper studies that tradeoff through \system{}. The same typed graph
representation, automatic differentiation, compiler, shader generator, memory
planner, checkpoint format, and runtime support inference and
optimizer-backed training. Vulkan is used on Linux, Windows, and Android, and
Metal on Apple platforms. The deployed application requires neither CUDA nor
ROCm nor a Python runtime.

The intended setting is a native application, robot, creative tool, or edge
device with mostly static tensor shapes and a reason to keep inference,
fine-tuning, or personalization in one codebase. \system{} is not positioned
as a replacement for dynamic Python research workflows, distributed
cloud-scale training, or the full operator breadth of mature frameworks.
Within that setting, the measured answer is more positive than the
conventional wisdom about graphics APIs suggests: both AMD devices reach
near-parity or better with compiled ROCm PyTorch on most workloads,
minimal-batch latency favors the portable stack on most devices, and the
only correctness-gate failures in fifty audited cells are two modes of one
APU workload whose cross-device gradients strongly implicate the reference
path. The gaps that
remain --- chiefly backward passes on NVIDIA and Apple hardware --- are
profiled down to the responsible kernel families, and the paper reports
them with the same prominence as the wins.

A further motivation is temporal. Deployed models today are frozen at
export: the training loop lives in a datacenter, the shipped artifact cannot
learn, and adaptation means a round trip through a separate stack ---
whereas biological learners adapt in place, continuously. Colocating a real
training loop with inference on the deployed device is a precondition for
studying that kind of live adaptation, and a shared train/deploy compiler is
the substrate it requires. This paper builds and measures the substrate;
live-adaptation behavior on top of it is deliberately left to follow-up
work.

We ask three research questions:
\begin{enumerate}
  \item \textbf{Performance portability:} how close does one implementation
        come to PyTorch for inference and forward--loss--backward execution
        across workloads, arithmetic policies, vendors, and device classes?
  \item \textbf{Causes:} which specializations recover performance, and which
        kernels, launches, or arithmetic constraints explain the remaining
        gaps?
  \item \textbf{Systems cost:} what does a shared training/deployment stack
        change about the artifact, application integration, correctness work,
        optimization strategy, and compiler authoring experience?
\end{enumerate}

This work makes four contributions:
\begin{itemize}
  \item A compact executable system whose shared graph/compiler path supports
        inference, reverse-mode differentiation, SGD and Adam updates, and
        checkpoint transfer into a fresh inference session.
  \item A controlled comparison with five matched model families, explicit
        arithmetic contracts, raw timing samples, and independent forward and
        backward correctness gates, frozen at one revision pair across five
        devices from three GPU vendors and Apple silicon --- 48 of 50 cells
        passing both gates, with cross-device gradient records localizing the
        two exceptions to the newly enabled reference path, subject to the
        need for a third implementation.
  \item A decomposition of performance into the kernel coverage, dispatch
        structure, cooperative-matrix use, and precision policy that produce
        both near-reference cases and the largest remaining gaps.
  \item Measured engineering results for deployment closure --- including a
        physical Android XR train-to-deploy case study --- automatic static
        execution, rewrite strategies, shader authoring boundaries, and their
        present debugging tradeoffs.
\end{itemize}

\section{Scope and Design Goals}

\paragraph{One stack for two phases.}
Inference and training are modes of one compiler rather than separate
products. Training differentiates the optimized forward graph, preserves the
forward outputs, appends one gradient output per parameter, and applies an SGD
or Adam update in the same static session. Parameters and optimizer state can
be stored in a safetensors checkpoint. A separately compiled inference session
with the same named parameters can load that checkpoint; it omits derivative
and optimizer nodes and may apply inference-only fusions. The performance
tables time forward--loss--backward separately from the optimizer so that
kernel/compiler cost is comparable, while Section~\ref{sec:train-deploy}
tests the complete workflow.

\paragraph{Portable execution, specialized programs.}
Portability does not imply one untuned kernel. \system{} generates specialized
programs for matrix shape, transpose mode, tile geometry, input format,
prologue, and epilogue. The invariant is that specialization remains behind
the same graph/runtime interface and lowers through the graphics stack.

\paragraph{Correctness before peak numbers.}
Every reported cell is conditional on a numerical contract. Inference and
backward execution are validated independently because an accelerated forward
pass can remain accurate while small derivative operands underflow in a
reduced input format.

\paragraph{Operational model and observability.}
Reducing host launch overhead in PyTorch can involve compiler integration or
CUDA Graph capture, whose warmup, static-shape, memory-address, and mutation
constraints are a recurring integration concern.\footnote{\href{
https://docs.pytorch.org/docs/stable/notes/cuda.html\#cuda-graphs}{
PyTorch CUDA Graphs documentation}.}
\system{} starts from a static graph: session construction fixes the dispatch
order, buffer ownership, pipelines, and synchronization, and each
\code{step()} executes that plan. There is no separate capture API or
application-managed graph-safe input buffer. This is an automation and
usability distinction, not a claim that Blade pre-records a native command
buffer or eliminates all host submission overhead.
The tradeoff is observability. PyTorch provides a mature eager mode,
operator-level inspection, autograd
diagnostics, and integrated profilers. \system{} currently provides graph and
dispatch-plan dumps, WGSL source-located parser errors, API validation,
per-dispatch GPU timings, gradient summaries, and Perfetto traces, but a
failure that crosses generated WGSL, Naga, Blade, and a vendor driver is still
hard to localize. Source mapping, automatic graph bisection, captured
intermediates, and a single operator replay tool are future work.

\paragraph{Scale of the implementation.}
``Compact'' is meant as a measurable property rather than a stylistic one.
At the frozen revision, the system is 34.5 KLOC of Rust spanning the typed
graph, automatic differentiation, rewrites, compiler, code generation, memory
planner, runtime, optimizer, ONNX/NNEF importers, and model builders, plus
6.1 KLOC of WGSL across 75 shader files; tests and examples add a further
14.8 KLOC.
These are physical source-line counts. That is the budget within which every
result below was produced. It is also
part of the explanation for them, in both directions: a kernel library of
this size cannot cover what a vendor stack covers, so the gaps analyzed in
Section~\ref{sec:gaps} are better read as a consequence of scope than as a
property of the graphics APIs --- while the cells at parity show what the
same budget already suffices for.

\paragraph{Non-goals.}
\system{} is not a complete replacement for PyTorch, does not cover arbitrary
dynamic Python programs or distributed training, and does not claim the
breadth of CUDA, cuDNN, or large production compilers. The evaluation asks
what this deliberately compact system achieves and where it stops.

\section{\system{} Architecture}

\begin{figure}[t]
\centering
\resizebox{\columnwidth}{!}{%
\begin{tikzpicture}[
  font=\sffamily\scriptsize,
  node distance=3.5mm,
  box/.style={
    draw=black!55,
    rounded corners=2pt,
    minimum height=7mm,
    align=center,
    inner xsep=4pt,
    inner ysep=2pt
  },
  input/.style={box, fill=gray!10},
  graph/.style={box, fill=blue!11},
  train/.style={box, fill=orange!15},
  compile/.style={box, fill=violet!11},
  runtime/.style={box, fill=green!13},
  backend/.style={box, fill=teal!13},
  arrow/.style={-{Latex[length=1.8mm]}, semithick, draw=black!65}
]
  \node[input, text width=31mm] (source)
    {Native builder\\or ONNX / NNEF importer};
  \node[graph, text width=26mm, below=of source] (typed)
    {Typed static graph};
  \node[graph, text width=26mm, below=of typed] (rewrite)
    {Graph rewrites and ordering};
  \node[graph, text width=23mm, below left=5mm and 3mm of rewrite] (infer)
    {Inference\\forward graph};
  \node[train, text width=27mm, below right=5mm and 3mm of rewrite] (training)
    {Autodiff\\forward + backward};
  \node[compile, text width=42mm, below=13mm of rewrite] (specialize)
    {Kernel archetypes + target specialization\\
     device capabilities + precision policy};
  \node[compile, text width=34mm, below=of specialize] (naga)
    {Generated WGSL\\validated Naga IR};
  \node[runtime, text width=34mm, below=of naga] (plan)
    {Static dispatch\\+ memory plan};
  \node[backend, text width=15mm, below left=4mm and 2mm of plan] (vulkan)
    {Vulkan};
  \node[backend, text width=15mm, below right=4mm and 2mm of plan] (metal)
    {Metal};

  \draw[arrow] (source) -- (typed);
  \draw[arrow] (typed) -- (rewrite);
  \draw[arrow] (rewrite.south) -- ++(0,-2mm) -| (infer.north);
  \draw[arrow] (rewrite.south) -- ++(0,-2mm) -| (training.north);
  \draw[arrow] (infer.south) |- (specialize.west);
  \draw[arrow] (training.south) |- (specialize.east);
  \draw[arrow] (specialize) -- (naga);
  \draw[arrow] (naga) -- (plan);
  \draw[arrow] (plan.south) -- ++(0,-2mm) -| (vulkan.north);
  \draw[arrow] (plan.south) -- ++(0,-2mm) -| (metal.north);

\end{tikzpicture}%
}
\caption{The shared training and inference compilation pipeline. Training adds
autodiff to the optimized forward graph; both modes then use the same
specialization, validation, scheduling, memory-planning, and backend path.}
\label{fig:pipeline}
\end{figure}
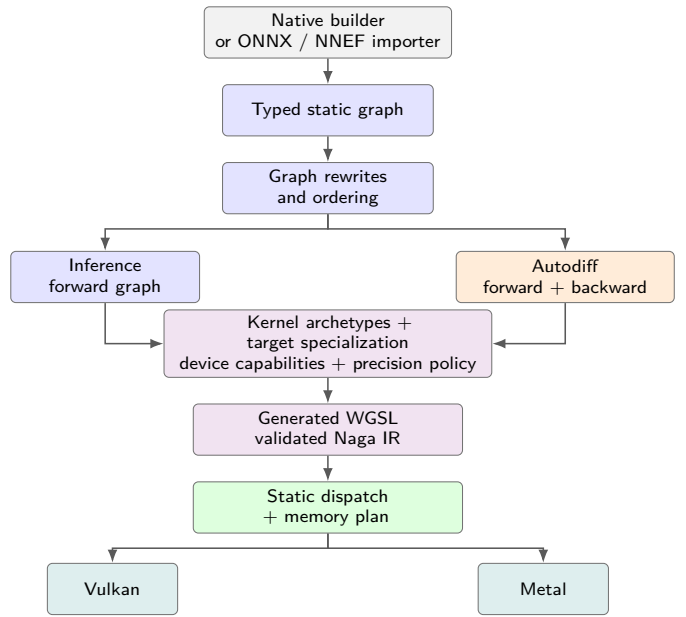

\subsection{Typed graph and automatic differentiation}

Graph nodes carry an operation, typed tensor shape, inputs, and a precision
policy. Builders cover dense and quantized parameters, pointwise operations,
reductions, normalization, matrix multiplication variants, convolution,
attention, indexing, and losses. ONNX and NNEF importers lower to the same
graph used by native model builders.

The node vocabulary is deliberately mid-level, and the choice is applied
consistently: an operation becomes a node when its differentiation rule is
local, its lowering maps to one kernel archetype or a short fusion of them,
and importers translate near one-to-one --- matrix products, convolutions,
attention, and normalizations are nodes; their scalar decompositions are
not. Both ends of the granularity spectrum carry real costs. Very low-level
primitive sets keep the IR minimal but push all performance recovery into
search: fused kernels must be rediscovered from clouds of primitives, which
is precisely the regime where global techniques such as equality saturation
become load-bearing. Monolithic high-level ops at the other end duplicate
kernel work per model family and starve the rewriter of reusable structure.
The mid-level choice keeps \system{}'s profitable rewrites local and
shallow --- one reason greedy rewriting suffices in
Section~\ref{sec:ablation} --- at the cost of a larger node vocabulary and
a per-node autodiff rule to maintain.

Training first optimizes and topologically sorts the forward graph, then
applies reverse-mode automatic differentiation. Multiple derivative paths are
combined explicitly in the graph. Backward-only nodes include transposed
matrix products, normalization derivatives, attention derivatives,
convolution input/weight derivatives, scatters, and shape-routing operations.

A precision bit is attached to derivative regions. It prevents an optional
f16-input cooperative-matrix promotion from rounding small gradient operands,
while still allowing a native f32 cooperative path where a device exposes
one. This policy was added after the Whisper workload showed that forward
agreement alone did not bound training error (Section~\ref{sec:precision}).

\subsection{Graph rewriting}

\system{} recognizes algebraic and implementation-level forms such as
$x\,\sigma(x)\rightarrow\mathrm{SiLU}(x)$, decomposed
SiLU-gating $\rightarrow$ SwiGLU, matrix-product plus residual, and packed
projection forms. A tensor-traffic cost estimates bytes read and written by an
extracted graph.

The compiler supports deterministic greedy rewriting and several
equality-saturation modes built with egglog~\cite{zhang2023egglog}. The latter
include fixed windows, repeated-region outlining, and whole-graph saturation.
Equality saturation is established prior art~\cite{willsey2021egg,
yang2021tensat}; our contribution is the measured result that it is not the
right production default for the current rewrite space. Section
\ref{sec:ablation} compares the modes.

\subsection{Kernel archetypes and specialization}

Rather than treating every operation as an unrelated shader, the code
generator uses a small conceptual set of archetypes:
\begin{itemize}
  \item pointwise DAGs with unary/binary inputs and chain fusion;
  \item workgroup reductions with composable per-element and output logic;
  \item matrix products with transpose modes, storage formats, prologues,
        epilogues, scalar tiles, GEMV, and cooperative tiles;
  \item implicit-GEMM and spatial convolution kernels and their derivatives;
  \item tiled attention forward, dQ, and fused dK+dV kernels.
\end{itemize}

Specialization is shape- and capability-aware. For example, a matrix
epilogue is part of the same pipeline/geometry key as its matrix kernel. A
cooperative implementation stages accumulator tiles through workgroup memory
before applying a scalar pointwise DAG. Weighted f16/Q4/Q8 kernels remain on
their own code-generation path until that path can compose with epilogues
without changing the B-buffer interpretation.

\subsection{WGSL, Naga, and graphics backends}

WebGPU is a standardized, safety-oriented graphics and compute API exposed by
browsers and native implementations. \code{wgpu} is a Rust implementation of
that API; Naga is its shader translation and validation library.\footnote{
\url{https://github.com/gfx-rs/wgpu/tree/trunk/naga}}
\system{} does not execute through a browser or \code{wgpu}. It uses Blade, a
smaller native graphics abstraction, to submit compute through Vulkan on
Linux, Windows, and Android and Metal on Apple platforms.\footnote{
\url{https://github.com/kvark/blade}}
Blade issues global barriers at pass boundaries instead of tracking
per-resource state; a companion cross-vendor study measures that model's
costs and headroom~\cite{malyshau2026barriers}.
Blade and \code{wgpu} nevertheless share Naga as a shader boundary, though
the integrations differ in two safety-relevant ways: Blade consumes the
validated module with Naga's runtime safety instrumentation disabled (no
injected bounds checks, unlike \code{wgpu}), and the binding model is
driven by Blade-side data declarations rather than reflected from the
shader. The shader boundary is validation and translation, not sandboxing
--- appropriate for a trusted, compiler-generated kernel set, and part of
what keeps the dispatch path thin.
This paper treats Blade as the device, resource, command, and presentation
substrate. \system{} owns the ML graph, autodiff, rewrites, kernel generation,
precision policy, dispatch/memory plan, optimizer, and evaluation studied
here; Blade's general graphics architecture is outside the paper's scope.

Generated or templated WGSL---the WebGPU shading language---is parsed and
validated by Naga, then consumed by Blade as a Naga module for SPIR-V/Vulkan
or Metal execution. WGSL is therefore an authoring and diagnostic boundary,
not evidence that the runtime itself is WebGPU and not a required source
round trip at execution time.

\system{} previously constructed Naga modules directly. This exposed arena
handles, expression-emission ranges, and errors such as ``Expression is not
cached'' without a useful source location. Immediately before retreating from
that design, the direct-IR code generator had reached 6,359 Rust lines.
Generated WGSL replaced it with 949 Rust lines and 1,402 WGSL lines, a roughly
63\% reduction for the corresponding implementation. Section
\ref{sec:naga} discusses this negative result.

\subsection{Static execution and memory planning}
\label{sec:memplan}

Compilation emits a fixed dispatch sequence with explicit logical buffers.
The runtime groups dispatches at dependency barriers, computes buffer
lifetimes, and aliases step-local intermediates with disjoint live ranges.
Parameters, graph outputs, gradients, and stateful buffers remain pinned.
Device-local storage is preferred on discrete GPUs; host-visible storage
remains available for unified-memory devices and diagnostics.

Calling \code{step()} walks this precompiled sequence without rediscovering
the graph or asking the application to capture it. Blade still creates an
encoder, records the selected compute passes, and submits work for the step;
the current implementation is therefore graph-static but not equivalent to
native CUDA Graph command-buffer replay.

Capabilities select scalar, small-tile, cooperative-matrix, generated
attention, and storage-format-specific pipelines. This selection also fixes
workgroup geometry and padding. Keeping these decisions atomic prevents a
failure mode in which a scalar epilogue pipeline is dispatched with
cooperative geometry.

\section{Evaluation Methodology}
\label{sec:method}

\inferena{} is the open-source cross-framework harness used for the
evaluation.\footnote{\url{https://github.com/kvark/inferena}} It constructs
matched inputs and objectives, invokes each framework runner, synchronizes
timing boundaries, retains raw samples, and records precision, environment,
revision, output, and gradient metadata. The harness does not supply kernels
to either engine. Every artifact identifies its exact Git revision alongside
the \system{} revision.

\paragraph{Reference system.}
PyTorch is the primary reference because the central experiment needs matched
forward, loss, and backward execution for every workload, in addition to
accelerated vendor backends. It is a much broader framework, and the
comparison does not imply feature equivalence. An inference-only runtime would
be informative for a different question but could not serve as the common
reference for the shared training/inference claim. Likewise, the footprint
comparison in Section~\ref{sec:footprint} measures deployment closure, not
equal functionality.

\subsection{Matched workloads}

Table~\ref{tab:workloads} summarizes the audited graphs. PyTorch and
\system{} use matching shapes, objectives, canonical parameter names, and
deterministic inputs. Physical matrix layouts are transposed during
initialization where required.

\begin{table*}[t]
\centering
\small
\begin{tabularx}{\textwidth}{l r X X}
\toprule
Workload & Parameters & Full shape & Objective and scope \\
\midrule
SmolLM2-135M & 135M &
batch 1, sequence 128 &
30-layer decoder-only transformer; token cross entropy \\
SmolVLA action expert~\cite{shukor2025smolvla} & 99.85M &
batch 1, 50 action tokens, 16 VLM tokens &
16-layer action expert; output MSE \\
Scaled Stable Diffusion 1.x U-Net~\cite{rombach2022ldm} & 10.93M &
batch 1, $4\times32\times32$ latent; context $77\times768$ &
timestep-conditioned convolutional U-Net with self/cross-attention; noise MSE \\
ResNet-50~\cite{he2016resnet} & 25.53M &
batch 4, $3\times224\times224$ &
inference-folded normalization representation; class cross entropy \\
Whisper-tiny encoder~\cite{radford2022whisper} & 8.21M &
batch 1, $80\times3000$ mel input &
four encoder layers, 1,500 output positions; output MSE \\
\bottomrule
\end{tabularx}
\caption{Audited workloads. The diffusion graph preserves the characteristic
latent, timestep, and text-conditioning paths, but reduces width, depth, and
block count and is not checkpoint-compatible with SD~1.5. Whisper omits the
decoder.}
\label{tab:workloads}
\end{table*}

\subsection{Arithmetic contracts}
\label{sec:precision}

\inferena{} defaults to the practical accelerated configuration; passing
\code{--strict} disables documented reduced-input matrix paths. The paper
reports both. Accelerated mode asks how fast each engine's normal
quality-validated hardware path is, while strict f32 is the controlled
arithmetic comparison. Accelerated mode is not a matched-format experiment.

\begin{table*}[t]
\centering
\small
\begin{tabularx}{\textwidth}{lXX}
\toprule
Property & Practical default & \code{--strict} \\
\midrule
Persistent tensors & f32 & f32 \\
PyTorch & TF32/high permitted & TF32 disabled \\
\system{} scalar path & f32 & f32 \\
\system{} eligible forward matrices & f16 input, f32 accumulate & f32 \\
\system{} backward matrices & f32 & f32 \\
Output & f32 & f32 \\
\bottomrule
\end{tabularx}
\caption{Arithmetic permissions. TF32 and IEEE f16 have different exponent
ranges; accelerated results are therefore reported separately.}
\label{tab:precision}
\end{table*}

PyTorch strict mode requests its highest float32 matmul precision and disables
CUDA matmul and cuDNN TF32. The practical default permits TF32.
\system{} strict mode disables f16 cooperative matrix and cooperative f16
attention. Its practical default permits eligible forward matrix,
convolution, and attention inputs to be rounded locally to f16, accumulates in
f32, and stores f32. Backward remains f32 unless a future experimental path
passes the same gradient gate.

This distinction repaired a concrete error. An earlier Whisper accelerated
run had small forward error but large gradient disagreement because derivative
operands were rounded before cooperative matrix multiplication. Marking the
automatically differentiated region full precision reduced the development
run's total-gradient relative error from approximately 22\% to 0.023\%. The
frozen accelerated runs confirm the repair: Whisper's total-gradient error
is 0.023\% on both discrete devices where the cooperative path engages
(Section~\ref{sec:accel-results}).

\subsection{Timing and validity}

Each series uses five untimed warmups and at least 20 retained samples.
We report median, interquartile range, minimum, maximum, and raw samples.
GPU execution is synchronized at timing boundaries.

Gap profiles are collected only after the ordinary series and never replace
its latency. Blade hardware timestamps are retained for repeated executions
with one compute pass per plan dispatch, together with the selected pipeline,
workgroup geometry, logical input/output bytes, and driver-reported executable
statistics where available. The artifact records the profiled wall time,
timestamped GPU sum, and their ratio to the ordinary grouped-pass median,
because pass-level instrumentation itself can be expensive. Vulkan intervals
include the inter-pass barrier before the following dispatch; Metal uses
compute-encoder boundary counter samples. We therefore use these profiles to
rank end-to-end dispatch costs, not as instruction-level kernel timings.

\textit{Inference} is one complete no-gradient forward pass.
\textit{Latency} is an agreed minimal shape, not an engine-specific shortcut.
\textit{Forward--loss--backward} includes those three phases but no optimizer
update. This isolates graph/compiler and kernel work from a choice of SGD or
Adam; Section~\ref{sec:train-deploy} separately verifies a real update and
checkpoint handoff. \textit{Compile} includes graph construction,
optimization, and GPU pipeline construction for the measured sessions; model
download and parameter upload are excluded or reported separately.

\paragraph{Memory.}
Each result also records GPU memory, and the two engines account for it
differently enough that every per-phase figure carries an explicit basis
label. \system{} reports what its execution plan physically allocates after
lifetime-based aliasing, together with the same plan's logical buffer total;
PyTorch reports its caching allocator's peak allocated and peak reserved
bytes. Those quantities are not interchangeable and are never placed in one
column. The cross-engine figure is per-process device memory, taken from
\code{VK\_EXT\_memory\_budget} or Metal's current allocated size for
\system{} and from NVML, \code{amd-smi}, or the MPS driver-allocated size
for PyTorch. It is the comparable one because it is reported by the driver
rather than by either engine: it includes context, pipeline, and staging
overhead that neither internal accounting observes, and being per-process it
is unaffected by an unrelated workload sharing the device. A backend that
cannot report a figure records it as absent rather than as zero.

Two sampling limits accompany every memory cell. \system{} samples at phase
boundaries rather than continuously within a step, so its per-process figure
bounds residency from below. PyTorch's phases share one allocator pool
without releasing it, so each peak includes residency established by earlier
phases.

To keep qualitative language falsifiable, a valid result is called
\textit{competitive} only when its median is no more than $2\times$ the
PyTorch median in the same device, workload, shape, and arithmetic mode.
Exact ratios are always reported; the threshold does not enter an aggregate
metric.

For SmolLM, the 128-token full forward is reported explicitly as
\textit{prefill}. The current one-token graph does not carry a KV cache and is
therefore labeled \textit{stateless one-token latency}, not decode latency.
This paper makes no decode claim; a decode result would require both engines
to use a matched cache layout and cache length.

PyTorch is the numerical reference. Forward validity requires matching output
shape, relative output L2 error below 1\%, and symmetric relative loss error
below 1\%. A forward--loss--backward cell additionally requires matching
canonical trainable parameter sets, total-gradient-norm error below 5\%, and
relative L2 error below 5\% over the vector of per-parameter gradient norms. A
backward failure invalidates that cell without discarding a valid inference
cell.

\subsection{Device and revision controls}

Every artifact records clean \system{} and \inferena{} revisions, GPU and
driver identifiers, OS, API/backend, toolchain versions, precision switches,
optimizer mode, input metadata, timing samples, output fingerprints, and
validation diagnostics. The frozen matrix (Table~\ref{tab:devices}) spans
five machines: NVIDIA and AMD discrete GPUs, an AMD APU, an Intel
integrated GPU, and Apple silicon. All 50 result artifacts --- five
devices, five workloads, two arithmetic modes --- carry the same clean
revision pair (\system{} \code{7561a64}, \inferena{} \code{7ca9c5c7}); no
historical cells are mixed in. An integrated or edge-class device is
included only if both runners complete the audited workloads: the Intel
machine qualifies because the PyTorch reference completes there on its
documented CPU fallback, which the tables label explicitly, and the AMD
APU qualifies because its reference completes every workload --- its
backward-validation failure on one workload is a reported result
(Section~\ref{sec:apu}), not an exclusion.

\begin{table*}[t]
\centering
\footnotesize
\setlength{\tabcolsep}{3pt}
\begin{tabular}{l l l l l}
\toprule
Device & Class & OS & \system{} backend & PyTorch build (backend, mode) \\
\midrule
NVIDIA GeForce RTX 5070 & discrete, 12\,GB & Linux & Vulkan, 595.71.05 & 2.13.0+cu130 (CUDA 13.0, compiled) \\
AMD Radeon RX 7900 XT (RADV) & discrete, 20\,GB & Linux & Vulkan, Mesa 26.0.3 & 2.10.0+rocm7.1 (ROCm 7.1, compiled) \\
AMD Radeon 780M (RADV) & integrated & Linux & Vulkan, Mesa 25.2.8 & 2.12.0+rocm7.14.0 (ROCm 7.14, compiled) \\
Intel Graphics (RPL-U) & integrated & Linux & Vulkan, Mesa 26.0.3 & 2.11.0+xpu (CPU, eager) \\
Apple M3 & unified SoC & macOS 15.7.3 & Metal & 2.11.0 (MPS, eager) \\
\bottomrule
\end{tabular}

\caption{The frozen device matrix. Each PyTorch build is the newest vendor
wheel that functioned on that machine, which is why versions differ.
``Compiled'' means the measured series ran under \code{torch.compile};
``eager'' means the compiler was unavailable there (unsupported for MPS in
the harness; Inductor fell back on the Intel machine's CPU device). On the
Intel machine the \code{+xpu} wheel installs but exposes no usable XPU
device, so its reference executes on the CPU.}
\label{tab:devices}
\end{table*}

\section{Results}
\label{sec:results}

\subsection{Functional train-to-deploy check}
\label{sec:train-deploy}

The performance protocol deliberately stops before an optimizer update. To
verify the broader systems claim, the artifact also contains a deterministic
executable check in \code{examples/train\_deploy.rs}. It compiles a two-class
linear model, performs 40 GPU SGD steps over eight separable points, saves the
parameters as a safetensors checkpoint, constructs a fresh inference-only
session, loads that checkpoint, and runs the same inputs without autodiff or
optimizer dispatches.

On the development RTX~5080, cross-entropy fell from 0.693147 to 0.026063 and
the reloaded inference session classified 8/8 points correctly. This is a
capability and checkpoint-compatibility test, not evidence of model quality or
generalization. Its value is that it exercises the complete
graph$\rightarrow$autodiff$\rightarrow$update$\rightarrow$checkpoint
$\rightarrow$fresh-inference path that the timing columns intentionally
decompose.


\subsection{Application-scale train-to-deploy case study: DinoVision}
\label{sec:dinovision}

\begin{figure}[t]
  \centering
  \includegraphics[width=0.82\columnwidth]{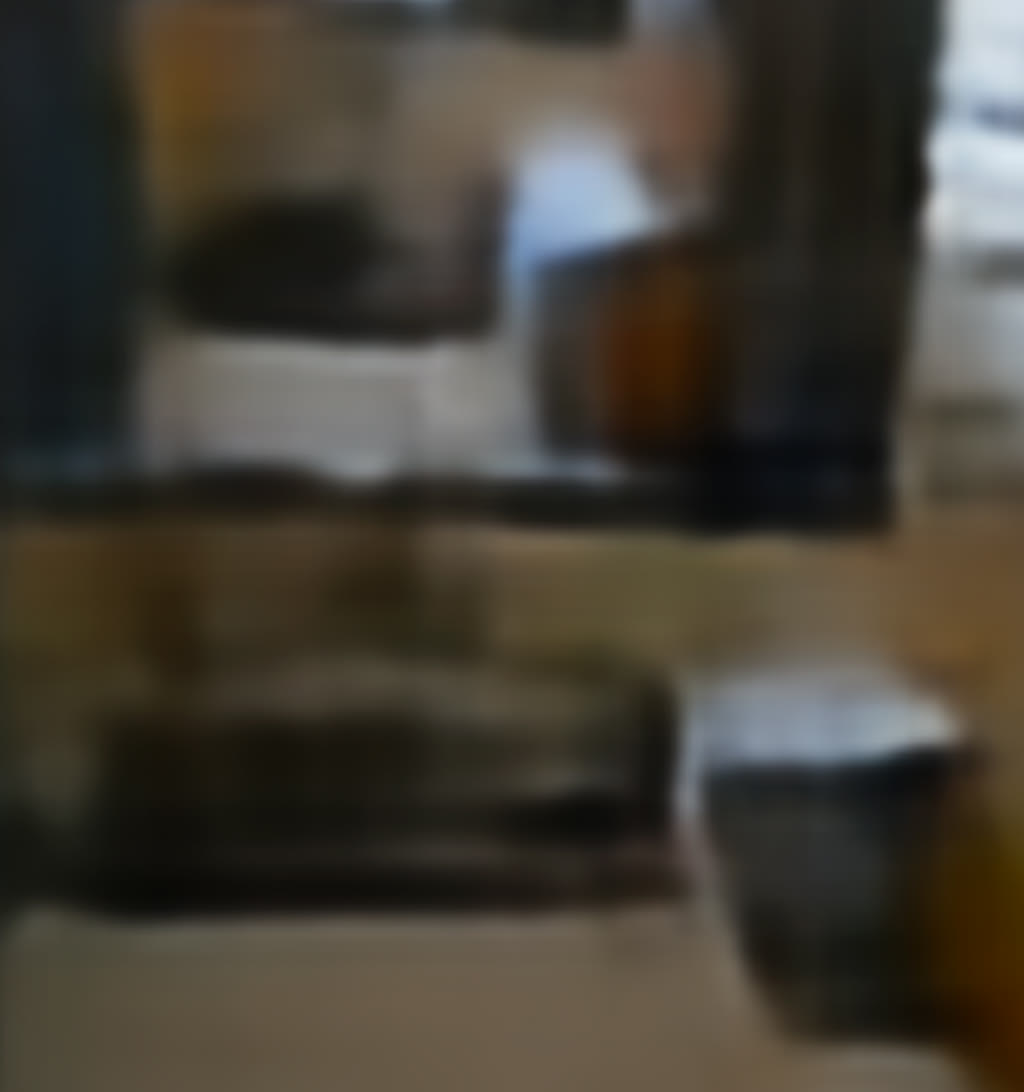}
  \caption{Quest 3S casting capture of DinoVision's live RGB reconstruction.
  This qualitative image documents physical deployment; the correctness and
  timing claims use the protocols described below.}
  \label{fig:dinovision-quest3s}
\end{figure}

We evaluate \system{} beyond isolated operators with DinoVision, an Android XR
application that reconstructs passthrough imagery from intermediate
DINOv3~\cite{simeoni2025dinov3} features
(Figure~\ref{fig:dinovision-quest3s}). The application center-crops and resizes
an RGB frame to $224\times224$, executes the first three layers of a frozen
DINOv3 ViT-S/16 encoder, rearranges its
$14\times14\times384$ patch features, and decodes them to RGB. On device, the
encoder and decoder form one batch-one inference graph, and two asynchronous
eye sessions share Blade's Vulkan context and queue with OpenXR rendering.

This case study exercises patch projection, learned prefix tokens, axial RoPE,
multi-head attention, LayerNorm, GELU MLPs, LayerScale, residual connections,
convolution, group normalization, SiLU, upsampling, autodiff, Adam, parameter
serialization, Android cross-compilation, and compute/graphics co-tenancy. It
is evidence about stack breadth and deployment closure, not a matched PyTorch
comparison or a capture-to-photon measurement.

\paragraph{Host training and held-out evaluation.}
The frozen encoder caches features on an NVIDIA host, after which \system{}
trains a 2,012,547-parameter decoder-only batch graph. Deployment uses the same
decoder construction and learned parameters in a joined batch-one graph. The
paths therefore share graph operators, compiler, memory planner, and runtime,
but are not literally the same complete graph.

We preserve Imagenette's upstream split. Three independently initialized
runs each use a class-balanced 2,500-image training subset, 12,000 batch-eight
Adam updates, and all 3,925 validation images. Seed zero was selected for
deployment before validation; seed one later scored highest. Across seeds,
global validation PSNR is $21.86\pm0.11$ dB, median-image PSNR is
$22.45\pm0.12$ dB, median RGB SSIM is $0.6405\pm0.0076$, and median MAE is
$0.04672\pm0.00070$ (mean $\pm$ sample standard deviation). The public
artifact retains every per-image record; host wall time is excluded because
the interactive machine experienced recorded suspension gaps.

\paragraph{Independent correctness gate.}
The independent gate proved necessary: an initial LayerScale layout mismatch
updated only the first token even though the output images remained plausible.
We rejected all affected weights and measurements, corrected the layout,
added exact attention and LayerScale CPU tests, and retrained all replicates.

An independent Torch/Transformers reference and \system{} consume the same
normalized f32 tensor and checkpoint. Predeclared thresholds are relative
$L_2\leq0.01$, CLS cosine $>0.999$, and every patch-token cosine $>0.999$.
Table~\ref{tab:dinovision-correctness} shows that the deployed and full-depth
controls pass.

\begin{table}[t]
  \centering
  \footnotesize
  \setlength{\tabcolsep}{4pt}
  \caption{\system{} encoder agreement with the independent reference.}
  \label{tab:dinovision-correctness}
  \begin{tabular}{rccc}
    \toprule
    Encoder depth & Relative $L_2$ & CLS cosine & Worst patch cosine \\
    \midrule
     1 & 0.000781 & 1.000000 & 0.999989 \\
     3 & 0.001403 & 1.000000 & 0.999995 \\
    12 & 0.002253 & 0.999998 & 0.999995 \\
    \bottomrule
  \end{tabular}
\end{table}

\paragraph{Physical Android correctness.}
Before timing, the host and Quest execute one public fixed RGB frame with
identical encoder and preselected decoder weights. Preprocessed patches are
bit-exact. Encoder output has relative $L_2=0.001059$, cosine $0.999999441$,
and minimum token cosine $0.999973147$; spatial decoder input has relative
$L_2=0.001063$ and cosine $0.999999438$; reconstruction has relative
$L_2=0.000457$ and cosine $0.999999938$.

\paragraph{Submission chunking under graphics co-tenancy.}
A native benchmark first measures the joined 2.359-GMAC graph in isolation.
Each chunk cell retains 20 synchronized samples after five warmups in three
fresh processes. We then run a live-worn stereo OpenXR sweep for 90 seconds
per cell at inference interval zero. Both sweeps use the predeclared order
$4,16,1,12,2,8$ to expose time drift. Table~\ref{tab:dinovision-chunks}
reports application render submissions, not compositor/display rate; worker
latency is not capture-to-photon latency. After discarding two initial
five-second windows, every cell retains 16 windows. All state snapshots report
an awake, worn headset and thermal status zero; peak reported GPU temperature
is $63.9\,^{\circ}$C.

\begin{table*}[t]
  \centering
  \caption{Isolated submission cost and live Blade/OpenXR co-tenancy on Quest
  3S. Brackets are interquartile ranges; overhead is relative to one chunk.}
  \label{tab:dinovision-chunks}
  \begin{tabular}{rrrrr}
    \toprule
    Chunks & Isolated ms [IQR] & Overhead & Render submit Hz &
    Lower-eye Hz / worker ms [IQR] \\
    \midrule
     1 & 96.78 [96.02, 97.41] & baseline & 10.83 & 5.99 / 149.51 [148.37, 150.99] \\
     2 & 100.79 [99.78, 101.46] & +4.1\% & 12.87 & 6.51 / 119.25 [118.73, 119.84] \\
     4 & 105.76 [104.75, 107.49] & +9.3\% & 12.07 & 6.06 / 129.69 [128.11, 142.42] \\
     8 & 108.25 [107.33, 109.99] & +11.9\% & 12.70 & 6.37 / 128.94 [127.26, 130.13] \\
    12 & 112.91 [112.06, 114.43] & +16.7\% & 12.61 & 6.33 / 131.57 [126.65, 134.12] \\
    16 & 114.68 [113.67, 115.48] & +18.5\% & 11.87 & 5.94 / 143.37 [140.24, 145.27] \\
    \bottomrule
  \end{tabular}
\end{table*}

Chunking incurs monotonic isolated cost, reaching 18.5\% at 16 chunks. Under
live co-tenancy, two chunks instead improve render submissions by 18.8\% and
the lower-eye update rate by 8.7\%, while reducing median worker latency by
20.2\% relative to one chunk. Larger counts regress non-monotonically, so the
result establishes chunking as a measurable, caller-selected scheduling
control rather than a universal two-chunk optimum. The current runtime
partitions ordered compiler barrier groups evenly by count; it does not
estimate their duration or observe renderer latency.

The experiment does not measure on-device training, PyTorch-on-Android
speedup, quantitative headset-camera quality, or capture-to-photon latency.
The current path includes CPU camera conversion and patchification, explicit
GPU transfers, output readback, temporal smoothing, and renderer upload.
Full weights, raw samples, manifests, environment records, and validation
scripts are available at
\url{https://huggingface.co/mad-bot/dinovision}; source is available at
\url{https://github.com/kvark/dinovision}. The artifact also provides the
full-resolution casting captures and a 39-second live video.

\subsection{Strict-f32 results}
\label{sec:strict-results}

\begin{table*}[!t]
\centering
\scriptsize
\setlength{\tabcolsep}{3.6pt}
\begin{tabular}{l rr rrr rrr rrr}
\toprule
 & \multicolumn{2}{c}{Compile (s)} & \multicolumn{3}{c}{Full / prefill (ms)} & \multicolumn{3}{c}{Minimal / one-token (ms)} & \multicolumn{3}{c}{F+L+B (ms)} \\
\cmidrule(lr){2-3}\cmidrule(lr){4-6}\cmidrule(lr){7-9}\cmidrule(lr){10-12}
Workload & Ours & PT & Ours & PT & $\times$ & Ours & PT & $\times$ & Ours & PT & $\times$ \\
\midrule
\multicolumn{12}{l}{\textbf{NVIDIA GeForce RTX 5070} --- \system{} Vulkan vs.\ PyTorch CUDA, compiled} \\
SmolLM2-135M & 1.52 & 50.2 & 12.41 & 6.44 & 1.93 & 2.89 & 3.21 & \textbf{0.90} & 47.00 & 16.35 & 2.87 \\
SmolVLA & 1.20 & 23.7 & 4.46 & 2.43 & 1.84 & 1.81 & 1.49 & 1.21 & 13.10 & 5.78 & 2.27 \\
Diffusion U-Net & 0.65 & 28.8 & 2.69 & 3.27 & \textbf{0.82} & 2.73 & 3.29 & \textbf{0.83} & 10.59 & 5.94 & 1.78 \\
ResNet-50 & 0.99 & 13.1 & 7.26 & 8.06 & \textbf{0.90} & 4.67 & 4.90 & \textbf{0.95} & 36.63 & 16.33 & 2.24 \\
Whisper-tiny & 0.74 & 7.3 & 9.04 & 3.41 & 2.65 & 9.03 & 3.42 & 2.64 & 33.27 & 11.97 & 2.78 \\
\midrule
\multicolumn{12}{l}{\textbf{AMD Radeon RX 7900 XT} --- \system{} Vulkan vs.\ PyTorch ROCm, compiled} \\
SmolLM2-135M & 0.47 & 96.0 & 10.79 & 9.78 & 1.10 & 1.93 & 6.65 & \textbf{0.29} & 39.80 & 32.68 & 1.22 \\
SmolVLA & 0.34 & 18.1 & 3.05 & 4.57 & \textbf{0.67} & 1.39 & 3.96 & \textbf{0.35} & 9.29 & 12.72 & \textbf{0.73} \\
Diffusion U-Net & 0.09 & 26.7 & 2.31 & 2.36 & \textbf{0.98} & 2.34 & 2.35 & \textbf{0.99} & 7.89 & 8.82 & \textbf{0.89} \\
ResNet-50 & 0.23 & 20.8 & 6.83 & 3.69 & 1.85 & 4.28 & 3.46 & 1.24 & 38.73 & 16.06 & 2.41 \\
Whisper-tiny & 0.21 & 6.7 & 6.75 & 6.34 & 1.06 & 6.74 & 6.49 & 1.04 & 25.04 & 27.91 & \textbf{0.90} \\
\midrule
\multicolumn{12}{l}{\textbf{AMD Radeon 780M} --- \system{} Vulkan vs.\ PyTorch ROCm, compiled} \\
SmolLM2-135M & 1.06 & 78.0 & 40.61 & 56.73 & \textbf{0.72} & 14.67 & 19.61 & \textbf{0.75} & 161 & 170 & \textbf{0.95} \\
SmolVLA & 0.76 & 38.7 & 16.03 & 22.40 & \textbf{0.72} & 11.00 & 14.07 & \textbf{0.78} & 54.68 & 58.37 & \textbf{0.94} \\
Diffusion U-Net & 0.25 & 50.5 & 11.60 & 5.59 & 2.07 & 11.56 & 4.64 & 2.49 & 20.10 & 12.41 & 1.62 \\
ResNet-50 & 0.77 & 28.1 & 76.53 & 41.52 & 1.84 & 22.30 & 10.87 & 2.05 & 211 & 131 & 1.60 \\
Whisper-tiny & 0.60 & 19.3 & 50.42 & 69.88 & \textbf{0.72} & 50.41 & 69.91 & \textbf{0.72} & 229 & 187 & 1.22$^\dagger$ \\
\midrule
\multicolumn{12}{l}{\textbf{Intel Graphics (RPL-U)} --- \system{} Vulkan vs.\ PyTorch CPU fallback, eager} \\
SmolLM2-135M & 1.19 & 0.0 & 148 & 201 & \textbf{0.74} & 22.97 & 36.84 & \textbf{0.62} & 835 & 832 & 1.00 \\
SmolVLA & 1.12 & 0.0 & 58.89 & 105 & \textbf{0.56} & 18.95 & 18.73 & 1.01 & 229 & 339 & \textbf{0.68} \\
Diffusion U-Net & 0.23 & 0.0 & 18.10 & 37.78 & \textbf{0.48} & 18.69 & 37.65 & \textbf{0.50} & 74.47 & 107 & \textbf{0.70} \\
ResNet-50 & 0.71 & 0.0 & 162 & 310 & \textbf{0.52} & 43.86 & 84.56 & \textbf{0.52} & 707 & 893 & \textbf{0.79} \\
Whisper-tiny & 0.61 & 0.0 & 187 & 307 & \textbf{0.61} & 188 & 291 & \textbf{0.65} & 1231 & 889 & 1.39 \\
\midrule
\multicolumn{12}{l}{\textbf{Apple M3} --- \system{} Metal vs.\ PyTorch MPS, eager} \\
SmolLM2-135M & 0.35 & 0.0 & 41.06 & 31.08 & 1.32 & 7.08 & 11.65 & \textbf{0.61} & 188 & 94.34 & 2.00 \\
SmolVLA & 0.24 & 0.0 & 19.95 & 10.63 & 1.88 & 5.24 & 8.21 & \textbf{0.64} & 59.73 & 29.06 & 2.06 \\
Diffusion U-Net & 0.08 & 0.0 & 6.76 & 7.29 & \textbf{0.93} & 6.76 & 7.60 & \textbf{0.89} & 24.56 & 20.47 & 1.20 \\
ResNet-50 & 0.19 & 0.0 & 49.94 & 34.36 & 1.45 & 15.97 & 10.30 & 1.55 & 216 & 92.34 & 2.34 \\
Whisper-tiny & 0.14 & 0.0 & 60.40 & 39.77 & 1.52 & 59.72 & 40.16 & 1.49 & 276 & 97.34 & 2.84 \\
\bottomrule
\end{tabular}

\caption{Strict-f32 results: medians over 20 samples after 5 warmups.
$\times$ is the \system{}/PyTorch ratio; bold marks cells where \system{} is
faster. F+L+B is forward, scalar loss, and backward without an optimizer
update. Every cell passes the forward gate of Section~\ref{sec:method}, and
every F+L+B cell except the marked one passes the backward gate.
$\dagger$: invalid --- the Whisper backward comparison on the 780M fails
the 5\% gradient gate; Section~\ref{sec:apu} presents cross-device evidence
implicating the reference path, but the ratio is shown for completeness only.
Whisper's
minimal shape equals its full shape, so its two forward columns coincide.
The Intel rows compare the integrated GPU against PyTorch's CPU fallback
and are reported as a support result, not an engine-efficiency result.}
\label{tab:strict-results}
\end{table*}

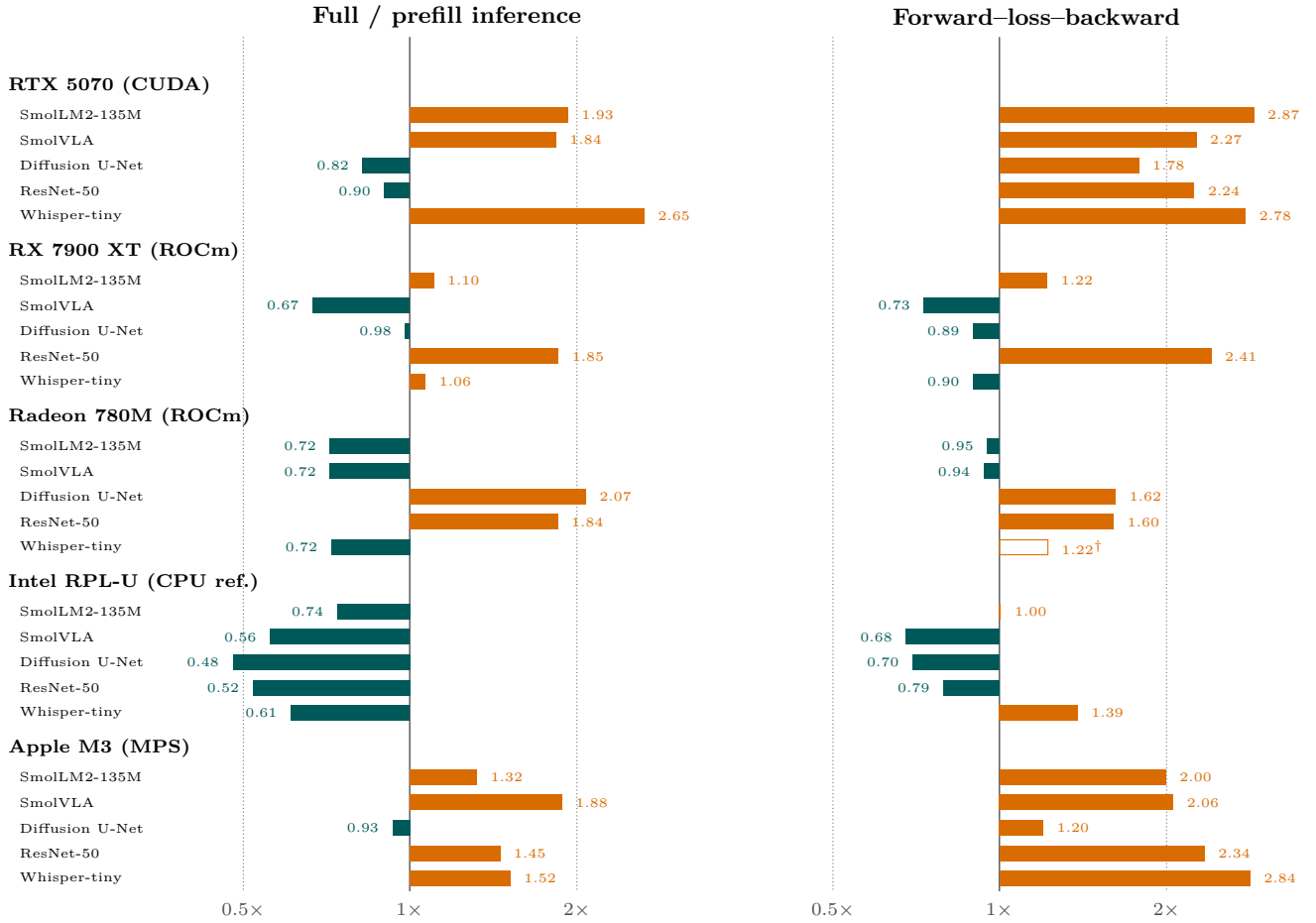
\begin{figure*}[!t]
\centering
\resizebox{0.98\textwidth}{!}{\begin{tikzpicture}[baseline]
\draw[densely dotted, black!60] (3.27,0.35) -- (3.27,-11.10);
\node[font=\scriptsize, black!70] at (3.27,-11.35) {0.5$\times$};
\draw[semithick, black!60] (5.50,0.35) -- (5.50,-11.10);
\node[font=\scriptsize, black!70] at (5.50,-11.35) {1$\times$};
\draw[densely dotted, black!60] (7.74,0.35) -- (7.74,-11.10);
\node[font=\scriptsize, black!70] at (7.74,-11.35) {2$\times$};
\node[font=\small\bfseries] at (6.00,0.62) {Full / prefill inference};
\draw[densely dotted, black!60] (11.17,0.35) -- (11.17,-11.10);
\node[font=\scriptsize, black!70] at (11.17,-11.35) {0.5$\times$};
\draw[semithick, black!60] (13.40,0.35) -- (13.40,-11.10);
\node[font=\scriptsize, black!70] at (13.40,-11.35) {1$\times$};
\draw[densely dotted, black!60] (15.64,0.35) -- (15.64,-11.10);
\node[font=\scriptsize, black!70] at (15.64,-11.35) {2$\times$};
\node[font=\small\bfseries] at (13.90,0.62) {Forward--loss--backward};
\node[anchor=west, font=\scriptsize\bfseries] at (0,-0.30) {RTX 5070 (CUDA)};
\node[anchor=west, font=\tiny] at (0.15,-0.69) {SmolLM2-135M};
\fill[orange!85!black] (5.50,-0.79) rectangle (7.62,-0.59);
\node[anchor=west, font=\tiny, orange!85!black] at (7.68,-0.69) {1.93};
\fill[orange!85!black] (13.40,-0.79) rectangle (16.81,-0.59);
\node[anchor=west, font=\tiny, orange!85!black] at (16.87,-0.69) {2.87};
\node[anchor=west, font=\tiny] at (0.15,-1.03) {SmolVLA};
\fill[orange!85!black] (5.50,-1.13) rectangle (7.46,-0.93);
\node[anchor=west, font=\tiny, orange!85!black] at (7.52,-1.03) {1.84};
\fill[orange!85!black] (13.40,-1.13) rectangle (16.04,-0.93);
\node[anchor=west, font=\tiny, orange!85!black] at (16.10,-1.03) {2.27};
\node[anchor=west, font=\tiny] at (0.15,-1.37) {Diffusion U-Net};
\fill[teal!70!black] (4.87,-1.47) rectangle (5.50,-1.27);
\node[anchor=east, font=\tiny, teal!70!black] at (4.81,-1.37) {0.82};
\fill[orange!85!black] (13.40,-1.47) rectangle (15.27,-1.27);
\node[anchor=west, font=\tiny, orange!85!black] at (15.33,-1.37) {1.78};
\node[anchor=west, font=\tiny] at (0.15,-1.71) {ResNet-50};
\fill[teal!70!black] (5.16,-1.81) rectangle (5.50,-1.61);
\node[anchor=east, font=\tiny, teal!70!black] at (5.10,-1.71) {0.90};
\fill[orange!85!black] (13.40,-1.81) rectangle (16.01,-1.61);
\node[anchor=west, font=\tiny, orange!85!black] at (16.07,-1.71) {2.24};
\node[anchor=west, font=\tiny] at (0.15,-2.05) {Whisper-tiny};
\fill[orange!85!black] (5.50,-2.15) rectangle (8.64,-1.95);
\node[anchor=west, font=\tiny, orange!85!black] at (8.70,-2.05) {2.65};
\fill[orange!85!black] (13.40,-2.15) rectangle (16.70,-1.95);
\node[anchor=west, font=\tiny, orange!85!black] at (16.76,-2.05) {2.78};
\node[anchor=west, font=\scriptsize\bfseries] at (0,-2.52) {RX 7900 XT (ROCm)};
\node[anchor=west, font=\tiny] at (0.15,-2.91) {SmolLM2-135M};
\fill[orange!85!black] (5.50,-3.01) rectangle (5.82,-2.81);
\node[anchor=west, font=\tiny, orange!85!black] at (5.88,-2.91) {1.10};
\fill[orange!85!black] (13.40,-3.01) rectangle (14.04,-2.81);
\node[anchor=west, font=\tiny, orange!85!black] at (14.10,-2.91) {1.22};
\node[anchor=west, font=\tiny] at (0.15,-3.25) {SmolVLA};
\fill[teal!70!black] (4.20,-3.35) rectangle (5.50,-3.15);
\node[anchor=east, font=\tiny, teal!70!black] at (4.14,-3.25) {0.67};
\fill[teal!70!black] (12.39,-3.35) rectangle (13.40,-3.15);
\node[anchor=east, font=\tiny, teal!70!black] at (12.33,-3.25) {0.73};
\node[anchor=west, font=\tiny] at (0.15,-3.59) {Diffusion U-Net};
\fill[teal!70!black] (5.44,-3.69) rectangle (5.50,-3.49);
\node[anchor=east, font=\tiny, teal!70!black] at (5.38,-3.59) {0.98};
\fill[teal!70!black] (13.05,-3.69) rectangle (13.40,-3.49);
\node[anchor=east, font=\tiny, teal!70!black] at (12.99,-3.59) {0.89};
\node[anchor=west, font=\tiny] at (0.15,-3.93) {ResNet-50};
\fill[orange!85!black] (5.50,-4.03) rectangle (7.49,-3.83);
\node[anchor=west, font=\tiny, orange!85!black] at (7.55,-3.93) {1.85};
\fill[orange!85!black] (13.40,-4.03) rectangle (16.24,-3.83);
\node[anchor=west, font=\tiny, orange!85!black] at (16.30,-3.93) {2.41};
\node[anchor=west, font=\tiny] at (0.15,-4.27) {Whisper-tiny};
\fill[orange!85!black] (5.50,-4.37) rectangle (5.71,-4.17);
\node[anchor=west, font=\tiny, orange!85!black] at (5.77,-4.27) {1.06};
\fill[teal!70!black] (13.05,-4.37) rectangle (13.40,-4.17);
\node[anchor=east, font=\tiny, teal!70!black] at (12.99,-4.27) {0.90};
\node[anchor=west, font=\scriptsize\bfseries] at (0,-4.74) {Radeon 780M (ROCm)};
\node[anchor=west, font=\tiny] at (0.15,-5.13) {SmolLM2-135M};
\fill[teal!70!black] (4.43,-5.23) rectangle (5.50,-5.03);
\node[anchor=east, font=\tiny, teal!70!black] at (4.37,-5.13) {0.72};
\fill[teal!70!black] (13.24,-5.23) rectangle (13.40,-5.03);
\node[anchor=east, font=\tiny, teal!70!black] at (13.18,-5.13) {0.95};
\node[anchor=west, font=\tiny] at (0.15,-5.47) {SmolVLA};
\fill[teal!70!black] (4.43,-5.57) rectangle (5.50,-5.37);
\node[anchor=east, font=\tiny, teal!70!black] at (4.37,-5.47) {0.72};
\fill[teal!70!black] (13.19,-5.57) rectangle (13.40,-5.37);
\node[anchor=east, font=\tiny, teal!70!black] at (13.13,-5.47) {0.94};
\node[anchor=west, font=\tiny] at (0.15,-5.81) {Diffusion U-Net};
\fill[orange!85!black] (5.50,-5.91) rectangle (7.86,-5.71);
\node[anchor=west, font=\tiny, orange!85!black] at (7.92,-5.81) {2.07};
\fill[orange!85!black] (13.40,-5.91) rectangle (14.96,-5.71);
\node[anchor=west, font=\tiny, orange!85!black] at (15.02,-5.81) {1.62};
\node[anchor=west, font=\tiny] at (0.15,-6.15) {ResNet-50};
\fill[orange!85!black] (5.50,-6.25) rectangle (7.48,-6.05);
\node[anchor=west, font=\tiny, orange!85!black] at (7.54,-6.15) {1.84};
\fill[orange!85!black] (13.40,-6.25) rectangle (14.93,-6.05);
\node[anchor=west, font=\tiny, orange!85!black] at (14.99,-6.15) {1.60};
\node[anchor=west, font=\tiny] at (0.15,-6.49) {Whisper-tiny};
\fill[teal!70!black] (4.45,-6.59) rectangle (5.50,-6.39);
\node[anchor=east, font=\tiny, teal!70!black] at (4.39,-6.49) {0.72};
\draw[orange!85!black] (13.40,-6.59) rectangle (14.05,-6.39);
\node[anchor=west, font=\tiny, orange!85!black] at (14.11,-6.49) {1.22$^\dagger$};
\node[anchor=west, font=\scriptsize\bfseries] at (0,-6.96) {Intel RPL-U (CPU ref.)};
\node[anchor=west, font=\tiny] at (0.15,-7.35) {SmolLM2-135M};
\fill[teal!70!black] (4.53,-7.45) rectangle (5.50,-7.25);
\node[anchor=east, font=\tiny, teal!70!black] at (4.47,-7.35) {0.74};
\fill[orange!85!black] (13.40,-7.45) rectangle (13.42,-7.25);
\node[anchor=west, font=\tiny, orange!85!black] at (13.48,-7.35) {1.00};
\node[anchor=west, font=\tiny] at (0.15,-7.69) {SmolVLA};
\fill[teal!70!black] (3.63,-7.79) rectangle (5.50,-7.59);
\node[anchor=east, font=\tiny, teal!70!black] at (3.57,-7.69) {0.56};
\fill[teal!70!black] (12.15,-7.79) rectangle (13.40,-7.59);
\node[anchor=east, font=\tiny, teal!70!black] at (12.09,-7.69) {0.68};
\node[anchor=west, font=\tiny] at (0.15,-8.03) {Diffusion U-Net};
\fill[teal!70!black] (3.13,-8.13) rectangle (5.50,-7.93);
\node[anchor=east, font=\tiny, teal!70!black] at (3.07,-8.03) {0.48};
\fill[teal!70!black] (12.24,-8.13) rectangle (13.40,-7.93);
\node[anchor=east, font=\tiny, teal!70!black] at (12.18,-8.03) {0.70};
\node[anchor=west, font=\tiny] at (0.15,-8.37) {ResNet-50};
\fill[teal!70!black] (3.40,-8.47) rectangle (5.50,-8.27);
\node[anchor=east, font=\tiny, teal!70!black] at (3.34,-8.37) {0.52};
\fill[teal!70!black] (12.65,-8.47) rectangle (13.40,-8.27);
\node[anchor=east, font=\tiny, teal!70!black] at (12.59,-8.37) {0.79};
\node[anchor=west, font=\tiny] at (0.15,-8.71) {Whisper-tiny};
\fill[teal!70!black] (3.91,-8.81) rectangle (5.50,-8.61);
\node[anchor=east, font=\tiny, teal!70!black] at (3.85,-8.71) {0.61};
\fill[orange!85!black] (13.40,-8.81) rectangle (14.45,-8.61);
\node[anchor=west, font=\tiny, orange!85!black] at (14.51,-8.71) {1.39};
\node[anchor=west, font=\scriptsize\bfseries] at (0,-9.18) {Apple M3 (MPS)};
\node[anchor=west, font=\tiny] at (0.15,-9.57) {SmolLM2-135M};
\fill[orange!85!black] (5.50,-9.67) rectangle (6.40,-9.47);
\node[anchor=west, font=\tiny, orange!85!black] at (6.46,-9.57) {1.32};
\fill[orange!85!black] (13.40,-9.67) rectangle (15.63,-9.47);
\node[anchor=west, font=\tiny, orange!85!black] at (15.69,-9.57) {2.00};
\node[anchor=west, font=\tiny] at (0.15,-9.91) {SmolVLA};
\fill[orange!85!black] (5.50,-10.01) rectangle (7.54,-9.81);
\node[anchor=west, font=\tiny, orange!85!black] at (7.60,-9.91) {1.88};
\fill[orange!85!black] (13.40,-10.01) rectangle (15.73,-9.81);
\node[anchor=west, font=\tiny, orange!85!black] at (15.79,-9.91) {2.06};
\node[anchor=west, font=\tiny] at (0.15,-10.25) {Diffusion U-Net};
\fill[teal!70!black] (5.27,-10.35) rectangle (5.50,-10.15);
\node[anchor=east, font=\tiny, teal!70!black] at (5.21,-10.25) {0.93};
\fill[orange!85!black] (13.40,-10.35) rectangle (13.99,-10.15);
\node[anchor=west, font=\tiny, orange!85!black] at (14.05,-10.25) {1.20};
\node[anchor=west, font=\tiny] at (0.15,-10.59) {ResNet-50};
\fill[orange!85!black] (5.50,-10.69) rectangle (6.71,-10.49);
\node[anchor=west, font=\tiny, orange!85!black] at (6.77,-10.59) {1.45};
\fill[orange!85!black] (13.40,-10.69) rectangle (16.15,-10.49);
\node[anchor=west, font=\tiny, orange!85!black] at (16.21,-10.59) {2.34};
\node[anchor=west, font=\tiny] at (0.15,-10.93) {Whisper-tiny};
\fill[orange!85!black] (5.50,-11.03) rectangle (6.85,-10.83);
\node[anchor=west, font=\tiny, orange!85!black] at (6.91,-10.93) {1.52};
\fill[orange!85!black] (13.40,-11.03) rectangle (16.76,-10.83);
\node[anchor=west, font=\tiny, orange!85!black] at (16.82,-10.93) {2.84};
\end{tikzpicture}}
\caption{Strict-f32 \system{}/PyTorch median ratios for all 25
device--workload cells (log scale; the solid line is parity). Teal bars
extend left of parity: \system{} is faster. Orange bars extend right:
PyTorch is faster. The hollow bar ($\dagger$) is the invalid 780M Whisper
backward cell of Section~\ref{sec:apu}. One-token latency, where \system{}
wins 12 of 20 GPU-referenced cells, is tabulated in
Table~\ref{tab:strict-results}. Generated from the same artifacts as the
tables.}
\label{fig:ratios}
\end{figure*}

Table~\ref{tab:strict-results} and Figure~\ref{fig:ratios} answer RQ1 with
raw per-device values rather than a universal-win claim, and the values
support more than parity-chasing.
Against \emph{compiled} CUDA PyTorch on the RTX~5070, \system{} is faster on
two of five inference cells (diffusion U-Net $0.82\times$, ResNet-50
$0.90\times$) and three of five one-token cells; the remaining inference
gaps are $1.84$--$2.65\times$, and training runs $1.8$--$2.9\times$ behind.
On the RX~7900~XT the portable stack effectively reaches the vendor stack:
four of five workloads are within $1.10\times$ for inference (SmolVLA is
$0.67\times$, i.e.\ 33\% faster) and \emph{three of five training cells are
outright faster} (SmolVLA $0.73\times$, diffusion U-Net $0.89\times$,
Whisper $0.90\times$), with ResNet-50 the outlier in both ($1.85\times$ and
$2.41\times$). On the Radeon~780M APU --- the machine whose vendor support
is newest --- \system{} wins three of five inference cells at $0.72\times$
and two training cells, with only the convolution pair beyond
$1.25\times$. On the Apple~M3, inference lands at $0.93$--$1.88\times$ and
training at $1.20$--$2.84\times$ against eager MPS --- the weakest surface
in the matrix, and the one that improved most from the profile-guided
optimization pass discussed in Section~\ref{sec:gaps}. On the Intel machine
the comparison inverts: the vendor XPU path does not function, and the
portable stack outperforms the resulting CPU reference in 12 of 15 cells
while being the only functional GPU path on that device.

Three cross-cutting observations. First, minimal-shape latency favors the
static dispatch plan: \system{} is outright faster in 12 of the 20
GPU-referenced one-token/minimal cells, most visibly on ROCm, where PyTorch's
one-token SmolLM step takes 6.65\,ms against 1.93\,ms ($3.4\times$). This
small shape is especially sensitive to launch and synchronization structure,
and the static dispatch sequence is faster despite \code{torch.compile}; the
measurement does not separate host launch cost from small-kernel quality. The
exceptions are
compute-bound minimal shapes (Whisper's full-length encoder, batch-1
ResNet-50 on the M3), where kernel quality decides instead. Second, the
convolution workloads are the only ones that resist parity on the AMD
devices: ResNet-50 training spans $1.60$--$2.41\times$ across the
GPU-referenced machines, while every non-convolution AMD training cell sits
at or below $1.22\times$ --- implicating convolution derivative coverage
rather than a uniform graphics-API tax. Section~\ref{sec:gaps} directly
confirms that kernel-family concentration for the largest NVIDIA gap.
Third, the correctness margins are wide: across all 25 strict cells the
worst forward relative L2 error is 0.0043\% against a 1\% gate, and across
the 24 valid backward comparisons the worst per-parameter gradient-norm
error is 0.64\% against a 5\% gate. Several per-cell error values reproduce to three digits across
all five devices, indicating that the residual disagreement is a
deterministic operation-ordering difference between the two engines rather
than platform noise.

\subsection{Accelerated arithmetic}
\label{sec:accel-results}

\begin{table*}[!t]
\centering
\scriptsize
\setlength{\tabcolsep}{4.2pt}
\begin{tabular}{l rrr rrr rrr}
\toprule
 & \multicolumn{3}{c}{Full / prefill (ms)} & \multicolumn{3}{c}{Minimal / one-token (ms)} & \multicolumn{3}{c}{F+L+B (ms)} \\
\cmidrule(lr){2-4}\cmidrule(lr){5-7}\cmidrule(lr){8-10}
Workload & Ours & PT & $\times$ & Ours & PT & $\times$ & Ours & PT & $\times$ \\
\midrule
\multicolumn{10}{l}{\textbf{NVIDIA GeForce RTX 5070} --- \system{} Vulkan vs.\ PyTorch CUDA, compiled} \\
SmolLM2-135M & 9.90 & 3.45 & 2.87 & 2.86 & 3.22 & \textbf{0.89} & 44.28 & 11.21 & 3.95 \\
SmolVLA & 4.45 & 1.72 & 2.59 & 1.80 & 1.45 & 1.24 & 13.02 & 5.75 & 2.26 \\
Diffusion U-Net & 2.18 & 1.72 & 1.27 & 2.28 & 1.74 & 1.31 & 10.26 & 6.08 & 1.69 \\
ResNet-50 & 7.29 & 2.64 & 2.76 & 4.67 & 1.60 & 2.91 & 36.77 & 7.94 & 4.63 \\
Whisper-tiny & 4.91 & 2.81 & 1.75 & 4.91 & 2.81 & 1.75 & 29.14 & 10.47 & 2.78 \\
\midrule
\multicolumn{10}{l}{\textbf{AMD Radeon RX 7900 XT} --- \system{} Vulkan vs.\ PyTorch ROCm, compiled} \\
SmolLM2-135M & 9.77 & 9.86 & \textbf{0.99} & 1.96 & 6.69 & \textbf{0.29} & 38.14 & 32.84 & 1.16 \\
SmolVLA & 3.82 & 4.56 & \textbf{0.84} & 1.38 & 3.96 & \textbf{0.35} & 9.59 & 12.71 & \textbf{0.75} \\
Diffusion U-Net & 2.13 & 2.36 & \textbf{0.90} & 2.10 & 2.36 & \textbf{0.89} & 7.84 & 8.77 & \textbf{0.89} \\
ResNet-50 & 6.62 & 3.68 & 1.80 & 4.23 & 3.46 & 1.22 & 38.62 & 16.18 & 2.39 \\
Whisper-tiny & 4.78 & 6.32 & \textbf{0.76} & 4.79 & 6.49 & \textbf{0.74} & 23.60 & 28.02 & \textbf{0.84} \\
\midrule
\multicolumn{10}{l}{\textbf{AMD Radeon 780M} --- \system{} Vulkan vs.\ PyTorch ROCm, compiled} \\
SmolLM2-135M & 31.42 & 56.65 & \textbf{0.55} & 14.69 & 19.45 & \textbf{0.76} & 151 & 169 & \textbf{0.90} \\
SmolVLA & 19.27 & 22.36 & \textbf{0.86} & 11.09 & 14.20 & \textbf{0.78} & 55.15 & 58.19 & \textbf{0.95} \\
Diffusion U-Net & 11.36 & 5.65 & 2.01 & 11.16 & 4.73 & 2.36 & 19.77 & 12.41 & 1.59 \\
ResNet-50 & 78.75 & 41.44 & 1.90 & 22.49 & 10.84 & 2.07 & 207 & 132 & 1.57 \\
Whisper-tiny & 52.41 & 69.73 & \textbf{0.75} & 52.34 & 69.86 & \textbf{0.75} & 231 & 186 & 1.24$^\dagger$ \\
\midrule
\multicolumn{10}{l}{\textbf{Intel Graphics (RPL-U)} --- \system{} Vulkan vs.\ PyTorch CPU fallback, eager} \\
SmolLM2-135M & 162 & 302 & \textbf{0.54} & 22.97 & 25.70 & \textbf{0.89} & 859 & 968 & \textbf{0.89} \\
SmolVLA & 59.02 & 101 & \textbf{0.59} & 18.91 & 25.24 & \textbf{0.75} & 222 & 323 & \textbf{0.69} \\
Diffusion U-Net & 18.25 & 36.21 & \textbf{0.50} & 18.73 & 31.28 & \textbf{0.60} & 74.24 & 84.64 & \textbf{0.88} \\
ResNet-50 & 169 & 283 & \textbf{0.60} & 43.93 & 80.91 & \textbf{0.54} & 731 & 853 & \textbf{0.86} \\
Whisper-tiny & 194 & 312 & \textbf{0.62} & 195 & 295 & \textbf{0.66} & 1269 & 912 & 1.39 \\
\midrule
\multicolumn{10}{l}{\textbf{Apple M3} --- \system{} Metal vs.\ PyTorch MPS, eager} \\
SmolLM2-135M & 41.22 & 31.09 & 1.33 & 7.49 & 11.81 & \textbf{0.63} & 189 & 94.54 & 2.00 \\
SmolVLA & 20.07 & 11.49 & 1.75 & 5.45 & 7.84 & \textbf{0.69} & 59.89 & 29.06 & 2.06 \\
Diffusion U-Net & 6.80 & 7.44 & \textbf{0.91} & 6.80 & 7.58 & \textbf{0.90} & 24.56 & 20.13 & 1.22 \\
ResNet-50 & 50.52 & 34.38 & 1.47 & 16.06 & 10.33 & 1.55 & 216 & 93.28 & 2.32 \\
Whisper-tiny & 58.55 & 39.58 & 1.48 & 58.51 & 39.74 & 1.47 & 274 & 97.34 & 2.82 \\
\bottomrule
\end{tabular}

\caption{Accelerated results under the same gates: PyTorch may use TF32,
\system{} may round eligible forward matrix inputs to f16 with f32
accumulation. The two fast paths are not format-equivalent, so this table is
reported separately from Table~\ref{tab:strict-results} rather than merged
with it. Validation matches the strict table: every cell passes except the
marked 780M Whisper backward comparison ($\dagger$,
Section~\ref{sec:apu}). Compile times match the strict table within 2\,s
and are omitted.}
\label{tab:accel-results}
\end{table*}

Table~\ref{tab:accel-results} compares each engine's documented fast path
under the same correctness gates. The instructive result is \emph{where}
each contract engages. TF32 changes PyTorch only on NVIDIA (ResNet-50
inference gains $3.1\times$, SmolLM and the diffusion U-Net about
$1.9\times$); its ROCm timings are unchanged and its MPS timings move only
within run-to-run variation, confirming the permission is CUDA-specific in
practice. \system{}'s f16 cooperative-matrix
path engages exactly where the driver reports the capability --- the
NVIDIA and both AMD GPUs (Whisper inference gains $1.8\times$ and
$1.4\times$ on the discrete cards; SmolLM gains $1.3\times$ on the APU,
putting it at $0.55\times$ the compiled reference) --- and is absent on the
M3 and the Intel iGPU, whose accelerated timings match strict within noise.
The net effect is that the NVIDIA gap widens (training up to $4.6\times$ on
ResNet-50, where TF32 convolutions compound the strict-mode conv-derivative
gap), while both AMD devices keep near-parity: on the RX~7900~XT, four of
five inference cells sit at or below $0.99\times$ and four of five valid
training cells between $0.75\times$ and $1.16\times$.

Two honest wrinkles. On the RX~7900~XT, SmolVLA \emph{regresses} from
$3.05$ to $3.82$\,ms when the cooperative path is permitted, and the same
regression reproduces on the 780M: the current selection treats an
available cooperative tile as always profitable, and for these shapes on
RADV it is not --- a per-shape cost model is the indicated fix. And
\system{}'s ResNet-50 and SmolVLA cells gain nothing on NVIDIA because
their hot matrix sites do not currently promote to cooperative geometry
(Section~\ref{sec:ablation}). Correctness margins remain wide in this mode:
among valid cells, worst forward L2 error 0.59\% (gate 1\%), worst
gradient-norm figures 0.18\% total and 0.64\% per-parameter (gate 5\%), and
the Whisper precision repair of Section~\ref{sec:precision} holds at
0.023\% total-gradient error on both engaged discrete devices.

\subsection{Compilation and startup}

Preparing a workload for execution --- graph construction, optimization,
autodiff, and GPU pipeline creation for the inference, one-token, and
training sessions --- takes \system{} 0.08--1.5\,s per workload in strict
mode and at most 2.4\,s with cooperative variants enabled, on every device
in the matrix. Where \code{torch.compile} functions it takes 6.5--96\,s for
the corresponding specializations (Table~\ref{tab:strict-results}), the
maximum being SmolLM on ROCm: 96\,s against 0.47\,s. This is not only a
development-loop difference: sub-second specialization is what makes
compile-on-device practical for the deployment scenarios of
Section~\ref{sec:implications}, where a shipped application cannot assume a
warm compiler cache or tolerate a minute of stall.

\subsection{Memory}
\label{sec:memory}

Artifact size is only one deployment cost; the memory a workload holds while
running decides whether it fits on a shared device at all. LlamaWeb treats
its memory reduction as a headline portability result for exactly this
reason~\cite{levine2026llamaweb}. Every result artifact therefore records,
per device, workload, and phase, the quantities defined in
Section~\ref{sec:method}: \system{}'s planned physical allocation, the same
plan's logical total, and per-process device memory for both engines.

\begin{table}[t]
\centering
\footnotesize
\setlength{\tabcolsep}{3.6pt}
\begin{tabular}{l rrr rrr}
\toprule
 & \multicolumn{3}{c}{Training (MiB)} & \multicolumn{3}{c}{Inference (MiB)} \\
\cmidrule(lr){2-4}\cmidrule(lr){5-7}
Workload & Logical & Phys. & Saved & Logical & Phys. & Saved \\
\midrule
SmolLM2-135M & 1974 & 1707 & 14\% & 875 & 742 & 15\% \\
SmolVLA & 1024 & 978 & 4\% & 596 & 563 & 6\% \\
Diffusion U-Net & 159 & 118 & 26\% & 121 & 84 & 30\% \\
ResNet-50 & 1640 & 861 & 48\% & 877 & 283 & 68\% \\
Whisper-tiny & 978 & 558 & 43\% & 287 & 61 & 79\% \\
\bottomrule
\end{tabular}

\caption{The memory-planner ablation, measured: logical buffer totals versus
physically allocated bytes after lifetime-based aliasing
(Section~\ref{sec:memplan}). Plans are shape-deterministic, so these values
vary by at most 1\% across backends; the artifact records all devices.
Neither engine's timed training session holds optimizer state.}
\label{tab:memory}
\end{table}

The recorded logical/physical pair turns the planner design of
Section~\ref{sec:memplan} into a measured ablation without a separate run
(Table~\ref{tab:memory}). Lifetime aliasing recovers 4--48\% of training
residency and 6--79\% of inference residency, and the spread confirms the
design intuition: the saving tracks how much of a graph is step-local
intermediate rather than pinned parameter or gradient state, peaking for the
activation-heavy ResNet-50 and Whisper graphs and nearly vanishing for the
parameter-dominated SmolVLA.

Against PyTorch's caching-allocator training peak on CUDA, the static plan
costs more on four of five workloads (e.g.\ ResNet-50 861 versus 498\,MiB,
Whisper 558 versus 275\,MiB) and less on one (diffusion U-Net 118 versus
164\,MiB). A dynamic allocator can retire each activation the moment backward
consumes it, while the current plan reserves its high-water configuration for
the whole step; the return is that a session performs no allocation at step
time and cannot fragment or fail at step $N$. The bases differ, so these
numbers are read side by side, never as one column.

The driver-reported per-process figure adds what internal accounting misses.
For SmolLM on the RTX~5070 the benchmark process peaks at 3{,}420\,MiB while
the plans of its three sessions sum to 3{,}165\,MiB: context, pipelines,
staging, and heap retention cost roughly 255\,MiB on that driver. The figure
is a whole-benchmark-process peak --- the harness keeps the inference and
one-token sessions resident together, and the allocator can retain heap from
a dropped session --- so it upper-bounds any single-session deployment. Even so it cuts
both ways by workload: for the diffusion U-Net on NVIDIA the whole \system{}
process peaks at 281\,MiB against PyTorch's 802\,MiB, and for Whisper on the
M3 at 618\,MiB against 1{,}236\,MiB, while for SmolLM the three-session
\system{} process is the larger one. The artifact retains every figure with
its basis label.

\subsection{Deployment footprint}
\label{sec:footprint}

The uniform stack also changes what an application must ship. Table
\ref{tab:footprint} reports a development measurement on Linux x86-64. The
\system{} entry is the stripped \inferena{} runner, which embeds the graph
compiler, autodiff, runtime, model builders, and shader sources in one
executable. It dynamically links ordinary OS libraries and relies on the
installed GPU driver, but has no Python, CUDA, or ROCm userspace dependency.
The PyTorch closure was computed recursively from active Python package
metadata rather than from the complete development virtual environment:
active \code{Requires-Dist} edges and requested extras were followed from
\code{torch}, and each installed file was counted once. The packages-only
row sums the \code{torch} and \code{triton} distributions.

\begin{table}[t]
\centering
\small
\begin{tabularx}{\columnwidth}{Xr}
\toprule
Linux x86-64 artifact & Installed size \\
\midrule
\system{} benchmark executable, stripped & 12.8 MiB \\
PyTorch + Triton packages only & 1.75 GiB \\
PyTorch CUDA runtime dependency closure & 4.66 GiB \\
\bottomrule
\end{tabularx}
\caption{Deployment footprint, measured on the Linux development machine.
Model weights, Python itself, ordinary OS libraries, and the GPU driver are
excluded from both sides. PyTorch is a much broader framework, so this is a
deployment-cost comparison rather than a feature-equivalence claim.}
\label{tab:footprint}
\end{table}

The roughly two-orders-of-magnitude difference is relevant to edge and
application embedding, but it is not a substitute for model coverage or
performance. TinyIREE similarly treats runtime and artifact size as
first-class deployment results, although it studies embedded inference
rather than portable GPU training \cite{liu2022tinyiree}.

\paragraph{What the closure costs in practice.}
Assembling the reference stack for this study was itself a measurement of
the fragmentation in Figure~\ref{fig:fragmentation}. Each machine needed a
different vendor package channel: an extra wheel index for CUDA
(\code{cu130}), an extra index \emph{plus} exact version pins for ROCm
(PyPI's default \code{torch} resolves to the CUDA build otherwise), and a
full index replacement for Intel's \code{+xpu} build. The resulting PyTorch
versions span 2.10--2.13 (Table~\ref{tab:devices}) because each vendor
channel lags differently. The outcomes also differ: on the Intel machine the
\code{+xpu} wheel installs cleanly yet exposes no usable device, so the
measured reference is its CPU fallback, and \code{torch.compile} was
exercised only on the CUDA and ROCm machines. The portable side of the same
experiment is one statically specialized binary per platform plus the
system's installed graphics driver; no per-vendor package source, version
pin, or compute runtime was involved on any of the machines. The Radeon
780M APU makes the asymmetry sharpest; Section~\ref{sec:apu} reports its
bring-up cost and the one workload responsible for both correctness failures
in the matrix, whose cross-device gradients implicate the reference path.

\subsection{Performance portability}
\label{sec:portability}

\begin{table}[t]
\centering
\small
\setlength{\tabcolsep}{4.5pt}
\begin{tabular}{l rr rr rr}
\toprule
 & \multicolumn{2}{c}{Inference} & \multicolumn{2}{c}{Minimal} & \multicolumn{2}{c}{F+L+B} \\
\cmidrule(lr){2-3}\cmidrule(lr){4-5}\cmidrule(lr){6-7}
Workload & Ours & PT & Ours & PT & Ours & PT \\
\midrule
SmolLM2-135M & 0.79 & 0.87 & 1.00 & 0.55 & 0.62 & 0.99 \\
SmolVLA & 0.74 & 0.75 & 0.96 & 0.65 & 0.68 & 0.85 \\
Diffusion U-Net & 0.82 & 0.78 & 0.77 & 0.79 & 0.76 & 0.90 \\
ResNet-50 & 0.70 & 0.83 & 0.73 & 0.84 & 0.52 & 0.95 \\
Whisper-tiny & 0.69 & 0.83 & 0.70 & 0.84 & 0$^\dagger$ & 0.98 \\
\midrule
\textit{mean} & 0.75 & 0.81 & 0.83 & 0.73 & 0.52 & 0.93 \\
\bottomrule
\end{tabular}

\caption{Pennycook performance portability over the frozen five-machine set,
strict mode. Application efficiency on each machine is measured against the
best \emph{valid} time either engine achieved there, so values are $\le 1$
and PyTorch's CPU fallback on the Intel machine counts as its own result on
that machine. $\dagger$: an invalid cell is not a result, so \system{}'s
Whisper training entry scores zero under the metric's discipline --- even
though Section~\ref{sec:apu} presents evidence implicating the reference
path. Accelerated-mode values are in the artifact; they shift the
column means by at most 0.05 and change no ordering.}
\label{tab:portability}
\end{table}

Table~\ref{tab:portability} aggregates the per-device results with the
Pennycook metric~\cite{pennycook2016metric}: application efficiency per
machine (relative to the best valid time observed on that machine), combined
by harmonic mean over the frozen set, with zero for a machine on which an
application produces no valid result. The metric's severity is visible in
one cell: \system{}'s single invalid backward comparison zeroes its entire
Whisper training score, cutting the training mean from 0.62 to 0.52 ---
even though the cross-device evidence implicates the reference path
(Section~\ref{sec:apu}). We apply the discipline to ourselves and report
the zero. The remaining structure is symmetric in an informative way. For
forward--loss--backward the incumbent is clearly more portable (mean 0.93
versus 0.52): vendor training libraries are good wherever they run. For
full-shape inference the two are close (0.81 versus 0.75). For
minimal-shape latency the portable stack is the more performance-portable
system (0.83 versus 0.73). The result is consistent with the static plan
reducing sensitivity to backend launch and synchronization costs, although
the end-to-end measurements also include kernel quality.

\subsection{The newest vendor path: bring-up and a localized anomaly}
\label{sec:apu}

The Radeon 780M machine is a full member of the frozen matrix, and it is
also the machine whose vendor support is newest --- ROCm reached this APU
class only recently, and standing the reference up required a current Linux
firmware payload and a PyTorch wheel from the separate ROCm~7.14 channel,
after which the runs still emit persistent rocSHMEM warnings. The portable
stack ran unchanged against the installed Mesa driver. That freshness shows
up in the matrix's one recurring correctness anomaly.

Whisper's forward pass on this machine validates at 0.0016\%, but its
backward comparison fails the gate at 17.8\% per-parameter
gradient-norm error, in both arithmetic modes --- and the recorded
artifacts localize the disagreement. \system{}'s per-parameter gradient
norms on this machine are identical, at recorded precision, to its own
norms on the RX~7900~XT, where they pass against that machine's ROCm
reference at 0.035\%; PyTorch's norms meanwhile differ by 16.3\% between
its own two ROCm machines. This cross-device evidence strongly implicates the
newly enabled reference path, and the independent forward/backward gates did
exactly what they were designed to do: an accurate forward pass did not
launder a backward disagreement. We report the cell as invalid --- the
protocol has no mechanism to bless our own result without a reference ---
and do not claim that these records alone prove which implementation is
correct. A third implementation is needed to resolve the anomaly; until then,
the Whisper row in Table~\ref{tab:portability} correctly scores a zero.

The rest of the machine's column is unremarkable in the best way: the
reference's per-process memory figure is recorded as absent per the
absent-not-zero rule of Section~\ref{sec:method}, and performance follows
the pattern of the other AMD device --- \system{} ahead on the
transformer-family workloads, behind about $2\times$ on the convolution
pair.

\section{Implications for Systems and Products}
\label{sec:implications}

\paragraph{The API is only one layer of the result.}
A strong result on one workload would not prove that Vulkan or Metal is
intrinsically as fast as CUDA, and a slow result would not prove the opposite.
The measured unit is the complete stack: graph representation, fusion and
specialization coverage, generated kernels, dispatch structure, driver
compiler, and arithmetic policy. The useful systems conclusion is more
specific. General consumer GPU APIs expose enough compute and matrix
capability for a compact compiler to be competitive on covered graphs; the
remaining distance can then be attributed to concrete missing work rather
than to ``portability'' as an indivisible tax.

\paragraph{A shared stack changes the product boundary.}
A native application can construct or import a graph, train or adapt it, save
the resulting parameters, and run an inference-only plan without introducing
a Python service or converting to a second model representation. That is
particularly relevant to robotics, creative applications, games, and private
on-device personalization, where the ML runtime must coexist with rendering,
sensors, and application logic. The compact artifact and Blade context-sharing
path make \system{} a plausible embeddable component, but productization still
depends on broader import/operator coverage, versioned checkpoints, stronger
diagnostics, and sustained-device profiling.

\paragraph{What an edge demonstration should prove.}
For a device such as Meta Quest, a more informative artifact than a standalone
kernel benchmark would collect a small amount of user-specific data, update a
compact model head on-device, checkpoint it, and immediately use it for
inference in the same application. A personalized hand/controller gesture
classifier is a bounded example: it exercises Android/Vulkan deployment and
the train-to-infer handoff without requiring a large generative model. Such a
result should report memory, sustained latency, thermal behavior, and
interference with the render loop. DinoVision
(Section~\ref{sec:dinovision}) evaluates the preceding host-to-edge boundary
on a physical Quest: a host-trained decoder is checkpointed and deployed
through the same compiler/runtime in an application that shares the graphics
queue. It does not update parameters on the headset, so on-device adaptation
and its memory budget remain future work.

\section{Ablations and Gap Analysis}
\label{sec:ablation}

\subsection{Greedy rewriting versus equality saturation}

The current rewrite set is small and mostly locally profitable. In CPU-only
ablation runs, greedy rewriting reaches essentially the same active node
counts as windowed or outlined egglog. Representative SmolLM inference
optimization took 0.089\,ms with greedy rewriting, 32.6\,ms with windowed
egglog, and 2.94\,ms with repeated-region outlining. Whole-graph saturation
took 56.2\,ms for inference and 7.43\,s for the differentiated graph.
End-to-end GPU time for the e-graph modes was within run-to-run variation of
greedy.

This is a negative result for a strong ``e-graphs make \system{} fast'' claim.
The rewrite rules provide small benefits, but equality saturation has no
demonstrated runtime advantage over their deterministic fixed point. We
therefore use greedy rewriting as the production default and retain egglog as
research infrastructure for future non-local alternatives. This complements
TENSAT and Glenside~\cite{yang2021tensat,smith2021glenside}: equality
saturation becomes compelling when the representation and rewrite set create
meaningful global choices, not merely because a graph is a tensor graph.

\subsection{Cooperative matrices and epilogues}

Two lessons about cooperative-matrix promotion came at measured cost. The
first is that pipeline variant, workgroup geometry, padding, and epilogue
must be selected atomically: an early path that combined cooperative
workgroup counts with a scalar epilogue pipeline could leave output rows
unwritten, and the safe replacement stages accumulator tiles through
workgroup memory as one selection unit (Section~\ref{sec:memplan}). The
second is that historical performance numbers are meaningless without
graph-level inspection: restoring the once-removed cooperative-epilogue
capability did not restore the SmolLM inference ratio associated with it,
because the packed-SwiGLU rewrite had meanwhile eliminated the
epilogue-bearing matmuls it applied to --- the frozen accelerated cell
stands at $2.87\times$, and the recovery now depends on cooperative
promotion at the current graph's sites, gated by the cost model that the
SmolVLA regression (Section~\ref{sec:accel-results}) shows is necessary.

\subsection{Where the remaining time goes}
\label{sec:gaps}

The largest frozen gaps are ResNet-50 training on NVIDIA ($4.6\times$
accelerated, $2.2\times$ strict), SmolLM training on NVIDIA ($4.0\times$
accelerated, $2.9\times$ strict), and Whisper training on the M3
($2.8\times$). For the first and third we captured per-dispatch profiles
with the mechanism of Section~\ref{sec:method}; both were taken at the
immediately preceding revision \code{b1405a3}, a disclosure that carries
different weight in each case.

\paragraph{Convolution derivatives own the ResNet gap.}
The NVIDIA ResNet-50 accelerated training profile is directly comparable to
the frozen cell: its unprofiled control median (36.81\,ms) matches the
frozen table (36.77\,ms) within 0.1\%. Of its timestamped GPU time,
backward spatial convolution takes 73.0\% and forward convolution another
16.3\% --- convolution is 89\% of the step. Three derivative shaders
account for 28.5\,ms of the 36.8\,ms step
(\code{Conv2dGradWeightGemmSmall} 12.6\,ms over just 12 dispatches,
\code{Conv2dGradInputGemm} 8.8\,ms, \code{Conv2dGradWeightGemm} 7.2\,ms);
matrix, normalization, pointwise, and data movement together are under
11\%. The weight-gradient ``small'' variant averaging over 1\,ms per
dispatch marks the concrete occupancy target. This confirms, at dispatch
granularity, that the worst gap in the matrix is a coverage-and-tuning
property of the current convolution-derivative implementations rather than a
uniform graphics-API overhead --- TF32 cuDNN convolutions on the reference
side then widen it.

\paragraph{Attention backward owned the Metal gap --- and paid for the fix.}
The M3 Whisper training profile predates the frozen revision deliberately:
captured at \code{b1405a3}, it showed backward attention consuming 58.3\%
of GPU time (\code{FlashGradKV} alone 141.9\,ms of a 330\,ms step,
\code{FlashGradQ} 57.5\,ms), with backward matrix products a distant second
at 14.2\%. That breakdown motivated the Metal optimization pass that landed
in the frozen revision \code{7561a64}, which cut the frozen cell to
274\,ms --- a $1.20\times$ improvement that moved Apple's worst training
ratio from $3.4\times$ to $2.8\times$. The remaining Apple gaps
($1.20$--$2.84\times$ training against an eager reference, with every
Apple inference cell at or below $1.88\times$) stay concentrated in the
backward kernel set, where no cooperative path engages.

\paragraph{Structure, not kernels, where \system{} already wins.}
The third cause is dispatch structure. The ResNet-50 inference profile
records 177 barrier groups over 198 dispatches --- nearly one barrier per
dispatch --- and our companion cross-vendor study of Blade's barrier model
quantifies the cost of redundant compute-pass barriers and the headroom
from eliding them~\cite{malyshau2026barriers}. Those measurements predate
Blade's newer global resource-access tracking, so we cite them as
demonstrated headroom rather than as a gain this stack currently banks.
Together with the frozen latency results
(Section~\ref{sec:strict-results}), the picture is consistent: where launch
and synchronization structure dominates, \system{}'s static plan already
wins; where a single kernel family dominates, coverage decides.
Cooperative-matrix promotion is the remaining
policy gap: the accelerated contract helps only where hot matmuls promote
(Whisper $1.8\times$ on NVIDIA, SmolLM $1.3\times$ on the APU), leaves
NVIDIA ResNet-50 and SmolVLA unimproved, and regresses SmolVLA on both
RADV devices --- the promotion decision needs a per-shape cost model, not
merely wider coverage. The improvement paths are therefore concrete and
falsifiable: unified convolution derivatives, Metal backward attention and
matrix tuning, barrier elision informed by the companion study, and
cost-model-gated cooperative promotion --- each an absent specialization or
policy, none an API limitation identified so far.

\section{Lessons from Direct Naga-IR Authoring}
\label{sec:naga}

Naga's internal representation is an arena-based, explicitly typed module
structure: each function owns arenas of expressions and statements,
cross-references are integer handles into those arenas, and expression
evaluation is scheduled by explicit emission ranges rather than implied by
tree shape. That design serves machine construction, validation, and
translation between WGSL, SPIR-V, and MSL well --- and it is an effective
portability boundary --- but we found it a poor \emph{human authoring}
boundary for this project. Direct construction required explicit types,
globals, bindings, arenas, expression handles, statements, and emission
ranges, and a missing range could surface during a later backend
transformation with no source-located diagnostic. Retiring direct
construction removed the 6,359 lines of IR-building Rust it had grown to,
replacing them with 949 lines of text generation plus 1,402 lines of WGSL
templates --- a net reduction of roughly 63\% for the corresponding
surface --- while making every generated kernel independently inspectable.

The temporary Naga-IR$\rightarrow$WGSL$\rightarrow$Naga-IR round trip was a
normalization workaround, not an architectural requirement. We eventually
fixed direct module submission, yet retained WGSL authoring because:
\begin{itemize}
  \item generated kernels remain recognizable and independently inspectable;
  \item Naga's parser owns expression-arena and emission invariants;
  \item diagnostics identify source locations;
  \item templates still express target constants, prologues, epilogues, and
        cooperative variants.
\end{itemize}

The current system still has 75 WGSL files and 6.1 KLOC, so retreating from
direct IR did not solve shader-family proliferation. During this work we
removed separate dK and dV shaders after making fused dK+dV the compiler
invariant. The longer-term target is one typed generator per major archetype,
with specialized kernels retained only when profiles justify them. The
broader lesson is that the best interchange IR and the best authoring IR need
not be the same representation.

\section{Related Work}

\paragraph{Portable ML compilation and runtimes.}
TVM established graph/operator optimization and learned scheduling for
heterogeneous targets~\cite{chen2018tvm}. Glow and MLIR demonstrate typed,
multi-level compiler architectures and static planning
\cite{rotem2018glow,lattner2021mlir}; Triton exposes a tile-level language for
high-performance deep-learning kernels~\cite{tillet2019triton}. RAF generates
and optimizes training graphs, including automatic differentiation and mixed
precision, for vendor GPU systems~\cite{yu2023raf}. \system{} is narrower in
framework breadth and execution substrate: it asks how far a compact native
Vulkan/Metal stack can take a shared inference and backward path. TinyIREE
makes compiled-artifact and runtime footprint a first-class result for
embedded inference~\cite{liu2022tinyiree}; \system{} measures the same
deployment concern for a graphics-API executable that also includes training.

Burn is a close open-source implementation analogue: it is a Rust
training-and-inference framework whose composable automatic differentiation
supports GPU backends including Vulkan, Metal, and WebGPU
\cite{simard2026burn}. Its broader framework and backend design establish
that portable Rust autodiff is not itself new. The distinction measured here
is one frozen native graphics-stack path evaluated across devices with matched
forward/backward gates, vendor references, compile and deployment costs, and
physical application co-tenancy.

TensorFlow.js demonstrated browser-resident training and inference through
WebGL~\cite{smilkov2019tensorflowjs}; WebLLM shows that WebGPU can retain a
substantial fraction of native LLM inference performance
\cite{ruan2024webllm}. LlamaWeb is the closest recent broad graphics-API
performance-portability study: it implements memory-efficient,
multi-precision browser inference and evaluates 10 models on 16 devices from
eight vendors~\cite{levine2026llamaweb}. A complementary WebGPU dispatch study
separates API, framework, and shader overhead across Vulkan and Metal
implementations~\cite{maczan2026dispatch}. \system{} instead studies native
Vulkan/Metal, several model families, and automatically differentiated
training as well as inference. We claim neither graphics-API machine learning
nor portable automatic differentiation in isolation as novel.

\paragraph{On-device deployment and training.}
ExecuTorch is the closest production analogue to the deployment half of this
work: a PyTorch program is exported once and executed across mobile,
embedded, and desktop backends without reimplementation, at very large
deployed scale~\cite{nachin2026executorch}. Its scope is inference. IREE
similarly compiles a model through MLIR into host scheduling logic and
device executables, and its Vulkan/SPIR-V path is the closest architectural
precedent for compiling machine learning to a portable graphics
API.\footnote{\url{https://github.com/iree-org/iree}} \system{} differs in
what crosses the deployment boundary rather than in the ambition to cross
it: the same graph, compiler, memory plan, and runtime also produce the
backward pass and the optimizer update, so adaptation can occur on the
deployed device instead of in a separate training stack. Research on
training within edge budgets approaches the same goal from the algorithm
side, co-designing quantized sparse updates for microcontroller-class
memory~\cite{lin2022ondevice}; \system{} retains full-precision reverse-mode
differentiation and instead asks how much a portable GPU stack can carry.

\paragraph{Tensor graph search.}
TASO searches verified graph substitutions~\cite{jia2019taso}. egg and egglog
provide reusable equality-saturation and fixpoint-reasoning infrastructure
\cite{willsey2021egg,zhang2023egglog}. TENSAT applies equality saturation and
global extraction to tensor graphs~\cite{yang2021tensat}, while Glenside uses
access patterns for low-level tensor rewriting~\cite{smith2021glenside}.
\system{} contributes neither a new e-graph data structure nor a pure tensor IR;
its useful result here is the production ablation against greedy rewriting.

\paragraph{Performance portability.}
Pennycook et al.\ argue that portability must be measured over an explicit
platform set and propose an aggregate that becomes zero when a platform is
unsupported~\cite{pennycook2016metric}. We follow that discipline by freezing
the device set, reporting per-device results, and separating arithmetic
contracts before considering an aggregate.

\section{Limitations and Threats to Validity}

Five workloads do not establish complete operator or model coverage. The
diffusion workload has the latent, timestep-conditioning, self-attention, and
77-token text cross-attention structure of Stable Diffusion 1.x, but uses
three reduced-width levels, one residual block per stage, and GELU in place of
GEGLU. It is not an SD~1.5 topology or checkpoint and excludes the text
encoder, VAE, and sampler. Whisper omits the decoder; ResNet represents folded
inference normalization rather than training-time batch statistics.
Synthetic deterministic inputs test computation but not task quality.

PyTorch is a moving reference whose compiler and libraries vary by version,
driver, and platform, and the frozen matrix makes that concrete: the four
machines run PyTorch 2.10--2.13 because each vendor wheel channel lags
differently, and \code{torch.compile} was exercised only on the CUDA and
ROCm machines. The eager-only Apple and Intel references make those ratios
generous to \system{}; we note that the conclusion least favorable to us on
Apple --- training $1.4$--$3.4\times$ behind --- survives the asymmetry,
since the handicapped reference still wins. The Intel machine's CPU
reference also shows up to $\sim$1.5$\times$ swings between arithmetic modes
whose switches are no-ops on CPU, so Intel ratios should be read as coarse.
Conversely, graphics drivers may optimize SPIR-V or Metal differently. Exact
revisions, raw samples, and environment metadata make the experiment
reproducible but do not eliminate this external validity threat.

Strict f32 controls arithmetic permissions, not operation order or bitwise
identity. Accelerated mode intentionally compares different fast formats and
must not be interpreted as format equivalence. Gradient-norm gates can miss
elementwise cancellation; the artifact should add sampled or full gradient
vector comparisons where memory permits. Two SPIR-V decorations bear on
this and are currently unexploited through Blade/Naga:
\code{NoContraction}, which would pin FMA contraction and tighten
cross-vendor reproducibility of the strict contract, and
\code{RelaxedPrecision}, a portable reduced-precision hint distinct from
explicit f16 storage; both are future work at the shader boundary.

The implementation has been optimized most heavily on NVIDIA hardware.
Results on AMD, Apple, and integrated devices are therefore both a
portability test and a maturity test --- which makes the AMD parity result
more surprising, not less. The frozen matrix contains one machine per
vendor and class, AMD excepted (one discrete and one integrated machine),
so per-device results should not be read as vendor-wide generalizations. The
frozen matrix contains no Windows device: the Vulkan path is the same code
there, but this paper makes no measured performance claim for it. Android is
represented by the DinoVision physical-device case study
(Section~\ref{sec:dinovision}), not by a matched PyTorch/\system{} matrix
cell; it therefore supports deployment and co-tenancy claims but contributes
no value to the paper's aggregate performance-portability metric.

Naga's internal IR validation and Vulkan's validation of emitted SPIR-V are
distinct boundaries. The currently pinned backend triggers
\code{VUID-StandaloneSpirv-None-10684}, an open upstream Naga/wgpu
explicit-layout issue that wgpu suppresses but Blade reports.\footnote{
\url{https://github.com/gfx-rs/wgpu/issues/7696}} Execution and numerical
checks pass on every device, but we do not treat those checks as proof of
fully valid SPIR-V; the diagnostic is retained as an explicit artifact
limitation until a verified upstream correction is consumed.

\section{Conclusion}

Can one portable GPU stack span training and deployment at useful
performance? On the evidence of the frozen matrix, yes --- with measured,
attributable exceptions. A typed graph, automatic differentiation,
specialized kernels, static execution, checkpointing, and lifetime-based
memory planning fit in 34.5 KLOC of Rust and 6.1 KLOC of WGSL; 48 of 50
audited cells pass both forward and backward gates across five consumer
devices. The stack deploys as a 13\,MiB binary that compiles workloads in
seconds. Where kernel
coverage matches a workload, the graphics-API path reaches a mature
vendor-native reference: training and inference near-parity or wins on both
AMD devices against compiled ROCm PyTorch, wins against compiled CUDA
PyTorch on two strict inference cells, and the stronger Pennycook
portability score for minimal-batch latency. On the machine where vendor
support arrived newest, the roles inverted: the portable stack ran
everything, and cross-device gradient records strongly implicate the
reference path in the matrix's only failed gates. Where coverage runs out,
the gap is profiled, not guessed: convolution derivatives are 89\% of the
worst step, Metal attention backward was 58\% of the second-worst before a
profile-guided pass reduced it, while cooperative-matrix coverage and
selection explain several arithmetic-policy shifts. These are concrete
specialization and policy targets; no API limitation is identified by the
measurements.

The shared stack does not eliminate systems work; it moves that work from
several conversion/runtime boundaries into one inspectable compiler and
artifact. The experience identifies the boundaries that currently scale:
precision policy must propagate through autodiff, pipeline geometry and
epilogues must be selected together, static execution can hide graph capture
from the application, greedy rewriting is more practical than equality
saturation for the present local rule set, and generated WGSL is more
maintainable than direct Naga-IR authoring. The result is therefore both a
usable implementation and a map of the remaining work required for portable
training and inference to become routine.

\section*{Artifact Availability}

Source code is available in the public \system{} and \inferena{}
repositories.\footnote{\url{https://github.com/kvark/meganeura}}\footnote{
\url{https://github.com/kvark/inferena}} Every benchmark-matrix table and the
ratio figure derive from the revision pair \system{} \code{7561a64} and
\inferena{} \code{7ca9c5c7} (both preserved under the public tag
\code{paper-arxiv-1}), recorded in each result file. The artifact includes the
raw JSON for all 50 cells --- timing
samples, correctness diagnostics, memory figures with basis labels, and
environment metadata --- plus the per-dispatch profile sidecars of
Section~\ref{sec:gaps} (which record their own revision), together with the
script that regenerates the benchmark-matrix tables and ratio figure from
those files.
The DinoVision case study (Section~\ref{sec:dinovision}) is its own frozen
artifact: source at \url{https://github.com/kvark/dinovision}, weights and
evidence records at \url{https://huggingface.co/mad-bot/dinovision}, with
its evidence revisions documented there. The repository includes
machine-readable citation metadata, and both benchmark revisions have public
frozen tags.

\section*{Acknowledgments}

The author thanks the wgpu community for maintaining Naga, the shader
validation and translation library that anchors \system{}'s portable shader
boundary, and thanks their family for support and patience throughout this
work.

\section*{AI-Assistance Disclosure}

Generative AI tools were used for code review, benchmark-harness development,
experimental debugging, and editorial assistance. The author designed the
study, reviewed generated changes, executed the experiments, verified the
reported claims and references, and accepts responsibility for the
manuscript.

\bibliographystyle{plain}
\bibliography{references}

@inproceedings{chen2018tvm,
  author = {Tianqi Chen and Thierry Moreau and Ziheng Jiang and Lianmin Zheng and Eddie Yan and Haichen Shen and Meghan Cowan and Leyuan Wang and Yuwei Hu and Luis Ceze and Carlos Guestrin and Arvind Krishnamurthy},
  title = {{TVM}: An Automated End-to-End Optimizing Compiler for Deep Learning},
  booktitle = {13th USENIX Symposium on Operating Systems Design and Implementation (OSDI 18)},
  year = {2018},
  pages = {578--594},
  publisher = {USENIX Association},
  url = {https://www.usenix.org/conference/osdi18/presentation/chen}
}

@article{rotem2018glow,
  author = {Nadav Rotem and Jordan Fix and Saleem Abdulrasool and Garret Catron and Summer Deng and Roman Dzhabarov and Nick Gibson and James Hegeman and Meghan Lele and Roman Levenstein and Jack Montgomery and Bert Maher and Satish Nadathur and Jakob Olesen and Jongsoo Park and Artem Rakhov and Misha Smelyanskiy},
  title = {Glow: Graph Lowering Compiler Techniques for Neural Networks},
  journal = {arXiv preprint arXiv:1805.00907},
  year = {2018},
  url = {https://arxiv.org/abs/1805.00907}
}

@inproceedings{lattner2021mlir,
  author = {Chris Lattner and Mehdi Amini and Uday Bondhugula and Albert Cohen and Andy Davis and Jacques Pienaar and River Riddle and Tatiana Shpeisman and Nicolas Vasilache and Oleksandr Zinenko},
  title = {{MLIR}: Scaling Compiler Infrastructure for Domain Specific Computation},
  booktitle = {2021 IEEE/ACM International Symposium on Code Generation and Optimization (CGO)},
  year = {2021},
  pages = {2--14},
  doi = {10.1109/CGO51591.2021.9370308}
}

@inproceedings{tillet2019triton,
  author = {Philippe Tillet and H. T. Kung and David Cox},
  title = {Triton: An Intermediate Language and Compiler for Tiled Neural Network Computations},
  booktitle = {Proceedings of the 3rd ACM SIGPLAN International Workshop on Machine Learning and Programming Languages},
  year = {2019},
  pages = {10--19},
  doi = {10.1145/3315508.3329973}
}

@article{yu2023raf,
  author = {Cody Hao Yu and Haozheng Fan and Guangtai Huang and Zhen Jia and Yizhi Liu and Jie Wang and Zach Zheng and Yuan Zhou and Haichen Shen and Junru Shao and Mu Li and Yida Wang},
  title = {{RAF}: Holistic Compilation for Deep Learning Model Training},
  journal = {arXiv preprint arXiv:2303.04759},
  year = {2023},
  url = {https://arxiv.org/abs/2303.04759}
}

@misc{simard2026burn,
  author = {Nathaniel Simard and Louis Fortier-Dubois and Dilshod Tadjibaev and Guillaume Lagrange and {Burn Framework Contributors}},
  title = {Burn},
  year = {2026},
  howpublished = {Software, version 0.21.0},
  url = {https://github.com/tracel-ai/burn}
}

@inproceedings{smilkov2019tensorflowjs,
  author = {Daniel Smilkov and Nikhil Thorat and Yannick Assogba and Ann Yuan and Nick Kreeger and Ping Yu and Kangyi Zhang and Shanqing Cai and Eric Nielsen and David Soergel and Stan Bileschi and Michael Terry and Charles Nicholson and Sandeep N. Gupta and Sarah Sirajuddin and D. Sculley and Rajat Monga and Greg Corrado and Fernanda B. Vi{\'e}gas and Martin Wattenberg},
  title = {{TensorFlow.js}: Machine Learning for the Web and Beyond},
  booktitle = {Proceedings of Machine Learning and Systems},
  volume = {1},
  year = {2019},
  url = {https://proceedings.mlsys.org/paper/2019/hash/acd593d2db87a799a8d3da5a860c028e-Abstract.html}
}

@article{levine2026llamaweb,
  author = {Reese Levine and Rithik Sharma and Nikhil Jain and Abhijit Ramesh and Zheyuan Chen and Neha Abbas and James Contini and Tyler Sorensen},
  title = {Llamas on the Web: Memory-Efficient, Performance-Portable, and Multi-Precision LLM Inference with {WebGPU}},
  journal = {arXiv preprint arXiv:2605.20706},
  year = {2026},
  url = {https://arxiv.org/abs/2605.20706}
}

@article{ruan2024webllm,
  author = {Charlie F. Ruan and Yucheng Qin and Akaash R. Parthasarathy and Xun Zhou and Ruihang Lai and Hongyi Jin and Yixin Dong and Bohan Hou and Meng-Shiun Yu and Yiyan Zhai and Sudeep Agarwal and Hangrui Cao and Siyuan Feng and Tianqi Chen},
  title = {{WebLLM}: A High-Performance In-Browser {LLM} Inference Engine},
  journal = {arXiv preprint arXiv:2412.15803},
  year = {2024},
  url = {https://arxiv.org/abs/2412.15803}
}

@article{maczan2026dispatch,
  author = {J{\k{e}}drzej Maczan},
  title = {Characterizing {WebGPU} Dispatch Overhead for {LLM} Inference Across Four {GPU} Vendors, Three Backends, and Three Browsers},
  journal = {arXiv preprint arXiv:2604.02344},
  year = {2026},
  url = {https://arxiv.org/abs/2604.02344}
}

@article{nachin2026executorch,
  author = {Mergen Nachin and Digant Desai and Sicheng Stephen Jia and others},
  title = {{ExecuTorch}: A Unified {PyTorch} Solution to Run {AI} Models On-Device},
  journal = {arXiv preprint arXiv:2605.08195},
  year = {2026},
  url = {https://arxiv.org/abs/2605.08195}
}

@inproceedings{lin2022ondevice,
  author = {Ji Lin and Ligeng Zhu and Wei-Ming Chen and Wei-Chen Wang and Chuang Gan and Song Han},
  title = {On-Device Training Under 256{KB} Memory},
  booktitle = {Advances in Neural Information Processing Systems},
  volume = {35},
  year = {2022},
  url = {https://arxiv.org/abs/2206.15472}
}

@article{liu2022tinyiree,
  author = {Hsin-I Cindy Liu and Marius Brehler and Mahesh Ravishankar and Nicolas Vasilache and Ben Vanik and Stella Laurenzo},
  title = {{TinyIREE}: An {ML} Execution Environment for Embedded Systems from Compilation to Deployment},
  journal = {IEEE Micro},
  volume = {42},
  number = {5},
  pages = {9--16},
  year = {2022},
  doi = {10.1109/MM.2022.3178068},
  url = {https://arxiv.org/abs/2205.14479}
}

@inproceedings{jia2019taso,
  author = {Zhihao Jia and Oded Padon and James Thomas and Todd Warszawski and Matei Zaharia and Alex Aiken},
  title = {{TASO}: Optimizing Deep Learning Computation with Automatic Generation of Graph Substitutions},
  booktitle = {Proceedings of the 27th ACM Symposium on Operating Systems Principles},
  year = {2019},
  pages = {47--62},
  doi = {10.1145/3341301.3359630}
}

@article{willsey2021egg,
  author = {Max Willsey and Chandrakana Nandi and Yisu Remy Wang and Oliver Flatt and Zachary Tatlock and Pavel Panchekha},
  title = {egg: Fast and Extensible Equality Saturation},
  journal = {Proceedings of the ACM on Programming Languages},
  volume = {5},
  number = {POPL},
  year = {2021},
  doi = {10.1145/3434304}
}

@inproceedings{zhang2023egglog,
  author = {Yihong Zhang and Yisu Remy Wang and Oliver Flatt and David Cao and Philip Zucker and Eli Rosenthal and Zachary Tatlock and Max Willsey},
  title = {Better Together: Unifying Datalog and Equality Saturation},
  booktitle = {Proceedings of the 44th ACM SIGPLAN Conference on Programming Language Design and Implementation},
  year = {2023},
  doi = {10.1145/3591239}
}

@inproceedings{yang2021tensat,
  author = {Yichen Yang and Phitchaya Phothilimthana and Yisu Wang and Max Willsey and Sudip Roy and Jacques Pienaar},
  title = {Equality Saturation for Tensor Graph Superoptimization},
  booktitle = {Proceedings of Machine Learning and Systems},
  volume = {3},
  year = {2021},
  url = {https://proceedings.mlsys.org/paper/2021/hash/cc427d934a7f6c0663e5923f49eba531-Abstract.html}
}

@inproceedings{smith2021glenside,
  author = {Gus Henry Smith and Andrew Liu and Steven Lyubomirsky and Scott Davidson and Joseph McMahan and Michael Taylor and Luis Ceze and Zachary Tatlock},
  title = {Pure Tensor Program Rewriting via Access Patterns},
  booktitle = {Proceedings of the 5th ACM SIGPLAN International Symposium on Machine Programming},
  year = {2021},
  doi = {10.1145/3460945.3464953}
}

@inproceedings{pennycook2016metric,
  author = {Simon J. Pennycook and Jason D. Sewall and Victor W. Lee},
  title = {A Metric for Performance Portability},
  booktitle = {Proceedings of the 7th International Workshop in Performance
               Modeling, Benchmarking and Simulation of High Performance
               Computer Systems (PMBS)},
  year = {2016},
  eprint = {1611.07409},
  archivePrefix = {arXiv}
}

@inproceedings{paszke2019pytorch,
  author = {Adam Paszke and Sam Gross and Francisco Massa and Adam Lerer and James Bradbury and Gregory Chanan and Trevor Killeen and Zeming Lin and Natalia Gimelshein and Luca Antiga and Alban Desmaison and Andreas Kopf and Edward Yang and Zachary DeVito and Martin Raison and Alykhan Tejani and Sasank Chilamkurthy and Benoit Steiner and Lu Fang and Junjie Bai and Soumith Chintala},
  title = {{PyTorch}: An Imperative Style, High-Performance Deep Learning Library},
  booktitle = {Advances in Neural Information Processing Systems},
  volume = {32},
  year = {2019},
  url = {https://papers.neurips.cc/paper/9015-pytorch-an-imperative-style-high-performance-deep-learning-library}
}

@inproceedings{he2016resnet,
  author = {Kaiming He and Xiangyu Zhang and Shaoqing Ren and Jian Sun},
  title = {Deep Residual Learning for Image Recognition},
  booktitle = {Proceedings of the IEEE Conference on Computer Vision and Pattern Recognition},
  year = {2016},
  pages = {770--778},
  url = {https://openaccess.thecvf.com/content_cvpr_2016/html/He_Deep_Residual_Learning_CVPR_2016_paper.html}
}

@inproceedings{rombach2022ldm,
  author = {Robin Rombach and Andreas Blattmann and Dominik Lorenz and Patrick Esser and Bj{\"o}rn Ommer},
  title = {High-Resolution Image Synthesis with Latent Diffusion Models},
  booktitle = {Proceedings of the IEEE/CVF Conference on Computer Vision and Pattern Recognition},
  year = {2022},
  pages = {10684--10695},
  url = {https://openaccess.thecvf.com/content/CVPR2022/html/Rombach_High-Resolution_Image_Synthesis_With_Latent_Diffusion_Models_CVPR_2022_paper.html}
}

@article{radford2022whisper,
  author = {Alec Radford and Jong Wook Kim and Tao Xu and Greg Brockman and Christine McLeavey and Ilya Sutskever},
  title = {Robust Speech Recognition via Large-Scale Weak Supervision},
  journal = {arXiv preprint arXiv:2212.04356},
  year = {2022},
  url = {https://arxiv.org/abs/2212.04356}
}

@article{shukor2025smolvla,
  author = {Mustafa Shukor and Dana Aubakirova and Francesco Capuano and Pepijn Kooijmans and Steven Palma and Adil Zouitine and Michel Aractingi and Caroline Pascal and Martino Russi and Andres Marafioti and Simon Alibert and Matthieu Cord and Thomas Wolf and Remi Cadene},
  title = {{SmolVLA}: A Vision-Language-Action Model for Affordable and Efficient Robotics},
  journal = {arXiv preprint arXiv:2506.01844},
  year = {2025},
  url = {https://arxiv.org/abs/2506.01844}
}

@misc{malyshau2026barriers,
  title  = {Global Pass Barriers Without Per-Resource {RHI} Tracking:
            A Cross-Vendor Study with {Blade}},
  author = {Dzmitry Malyshau},
  year   = {2026},
  eprint = {2607.26506},
  archivePrefix = {arXiv},
  primaryClass = {cs.GR},
  howpublished = {arXiv:2607.26506}
}

@article{simeoni2025dinov3,
  title         = {{DINOv3}},
  author        = {Sim{\'e}oni, Oriane and Vo, Huy V. and Seitzer, Maximilian and
                   Baldassarre, Federico and Oquab, Maxime and Jose, Cijo and
                   Khalidov, Vasil and Szafraniec, Marc and Yi, Seungeun and
                   Ramamonjisoa, Micha{\"e}l and Massa, Francisco and Haziza, Daniel and
                   Wehrstedt, Luca and Wang, Jianyuan and Darcet, Timoth{\'e}e and
                   Moutakanni, Th{\'e}o and Sentana, Leonel and Roberts, Claire and
                   Vedaldi, Andrea and Tolan, Jamie and Brandt, John and
                   Couprie, Camille and Mairal, Julien and J{\'e}gou, Herv{\'e} and
                   Labatut, Patrick and Bojanowski, Piotr},
  journal       = {arXiv preprint arXiv:2508.10104},
  year          = {2025},
  eprint        = {2508.10104},
  archivePrefix = {arXiv},
  primaryClass  = {cs.CV},
  doi           = {10.48550/arXiv.2508.10104}
}

\end{document}